\documentclass[12pt]{article}

\ifx\directlua\undefined\ifx\XeTeXcharclass\undefined
  \else\RequirePackage[no-math]{fontspec}[2017/03/31]\fi 
  \else\RequirePackage[no-math]{fontspec}[2017/03/31]\fi 
\usepackage{amsmath,amssymb,graphicx}
\usepackage[top=1in, bottom=1in, left=1in, right=1in]{geometry}
\usepackage{graphicx} 
\usepackage{setspace}
\usepackage{adjustbox}
\usepackage{tabularx}

\usepackage[T1]{fontenc}
\usepackage[utf8]{inputenc}
\usepackage{lmodern}
\usepackage{tipa} 
\usepackage{mathrsfs}

\usepackage{multirow}
\usepackage{url}
\usepackage[table]{xcolor}
\usepackage{hyperref}
\usepackage{booktabs}
\usepackage[
  backend=biber,
  style=unified,
  natbib,
  language=english,
  maxcitenames=3,
  maxbibnames=99
]{biblatex}
                
\usepackage{csquotes}
\usepackage{enumitem}
\setlist[itemize]{font=\normalsize, label=\textbullet, itemsep=0.4ex, topsep=0pt}

\DeclareFieldFormat{postnote}{\mknormrange{#1}}
\DeclareFieldFormat{multipostnote}{\mknormrange{#1}}

\usepackage[final]{changes}

\begin{document}
\sloppy

\begin{center}
    \large \textbf{Integrating Human Linguistic Insights into AI: Theory-Driven Representation for Multilingual Text-to-Speech}
\end{center}

\begin{flushleft}
\noindent

\textbf{Authors}: 

Cong Zhang\textsuperscript{1, *}, 
Huinan Zeng\textsuperscript{2}, 
Huang Liu\textsuperscript{3,$\dagger$},
Jiewen Zheng\textsuperscript{4}\\[1em]

\textbf{Affiliations}: 

\textsuperscript{1}Speech and Language Sciences, Newcastle University, Newcastle upon Tyne, United Kingdom

\textsuperscript{2}Faculty of Linguistics, Philology and Phonetics, University of Oxford, Oxford, United Kingdom

\textsuperscript{3}Tencent, Shenzhen, China

\textsuperscript{4}Shengrurenxin Ltd., Beijing, China\\[1em]

\textbf{Corresponding author details}:

Cong Zhang

Address: 
King George VI Building\\
Newcastle upon Tyne\\
NE1 4LF\\
United Kingdom\\

Email: cong.zhang@newcastle.ac.uk
\\[1em]

\begingroup
\renewcommand{\thefootnote}{\fnsymbol{footnote}}
\footnotetext[1]{Corresponding author}
\footnotetext[2]{This work was conducted while the author was at Shengrurenxin Ltd., Beijing, China}
\endgroup
\end{flushleft}

\pagebreak
\doublespacing

\section*{Abstract}
This paper explores the integration of human linguistic insights into multilingual text-to-speech (TTS) systems by evaluating the Featurally Underspecified Lexicon (FUL) as a theory-driven input representation. Unlike data-intensive end-to-end models, FUL offers a compact, interpretable feature set grounded in phonological principles, enabling scalable and equitable TTS development for low-resource languages. We provide a mapping from language-specific phones to FUL feature vectors via a SAMPA intermediate and incorporate these features into a modified FastSpeech architecture. Experiments were conducted to evaluate their ability to generate native, non-native, and code-mixed speech in English and Mandarin. We ran an experiment with a small dataset and one with a larger dataset, which showed that TTS with FUL features as input could produce intelligible native speech with as little as 8 hours of training data; with 100 hours of training data, intelligible speech could be generated for a language not present in the training data. The approach further supports code-mixed synthesis while preserving consistent timbre and interpretable phonetic control. These results highlight the potential of theory-driven representations for building efficient, scalable, and linguistically informed TTS systems, demonstrating that phonological features can function as both analytical tools and practical inputs for speech technology. 
\\[1em]
\textbf{Keywords}: text to speech, phonological feature, multilingual, featurally underspecified lexicon, low-resource language

\pagebreak

\section{Introduction}

In recent years, large language models have revolutionised AI-driven speech technologies by training on vast amounts of data, and text-to-speech (TTS) is not an exception. While these models have achieved impressive performance, their data-hungry nature raises concerns about equity and scalability -- especially for low-resource languages and dialects. In contrast, humans are remarkably good at identifying linguistic patterns with minimal input, an ability that has guided decades of linguistic research. Linguists have developed formal representational frameworks, such as distinctive feature systems, to capture phonological contrasts observed across many languages. While these frameworks are not exhaustive and remain the subject of ongoing debate, they provide a structured account of linguistic generalisations across languages. As speech technology continues to evolve, one promising direction lies in integrating this deep well of human linguistic knowledge into data-driven AI systems, enabling more efficient and equitable speech technologies. 

To illustrate this approach, we investigate whether the compact, theory-driven phonological feature system proposed in Featurally Underspecified Lexicon (FUL) \citep{LahiriReetz2002, Lahiri2010, lahiri_predicting_2018} can serve as a shared input representation for multilingual TTS. We focus on two languages from different language families and with different prosodic systems, Mandarin and English, providing a useful test case for evaluating cross-linguistic generalisability.

The present study explores the integration of theory-driven phonological
representations into multilingual text-to-speech (TTS) systems. Rather than aiming to provide a comprehensive solution for improving performance, we position this work as a proof-of-concept evaluation of whether the Featurally Underspecified Lexicon
(FUL) can serve as a viable input representation for TTS. Accordingly, the immediate objectives of this paper are as follows:

\begin{enumerate}[label=(\arabic*)]
  
  \item To assess the feasibility of using phonological features in multilingual TTS, focusing on whether FUL features can support intelligible speech synthesis across both seen and unseen language conditions with relatively limited training data.

  \item To explore whether speech synthesis using FUL features can provide insight into potential theoretical implications for the FUL framework.
  
  \item To develop and demonstrate a mapping from common language-specific phone symbols (English: ARPABET; Mandarin: Pinyin) to FUL features via SAMPA (English: SAMPA \citep{wellsSAMPAComputerReadable1997}; Mandarin: SAMPA-SC \citep{zhang_hanyu_2009}), ensuring compatibility with existing TTS systems and labelled data.
\end{enumerate}

In the long term, the broader aim of this line of research is to investigate whether phonological feature-based discrete representations, such as FUL, can support scalable and data-efficient multilingual TTS, including for low-resource languages. \deleted{The present study is limited to a proof-of-concept evaluation using two languages. Further work is required to assess the generalisability of this approach across a wider range of languages, including genuinely low-resource settings, and to more fully evaluate its theoretical implications.}

\section{Background}

\subsection{TTS Modelling}
\label{sec:tts_modelling}
TTS systems have evolved through several stages. Early systems relied on formant and concatenative syntheses, producing intelligible but often unnatural speech \citep{klatt1987review}. These rule-based systems used grapheme-to-phoneme (G2P) conversion and required relatively little recorded data. In the mid-2000s, statistical parametric speech synthesis (SPSS) emerged, initially with hidden Markov models (HMMs) \citep{tokuda2000speech} and later deep neural networks (DNNs) \citep{zen2013statistical}. These systems predicted acoustic parameters from richer linguistic features, typically needing 5-15 hours of clean data per high-quality voice.

End-to-end models such as Tacotron \citep{wang2017tacotron} and Tacotron 2 \citep{shen2018natural} introduced sequence-to-sequence mapping from input text or phonemes to mel-spectrograms, synthesised with neural vocoders. Successors like FastSpeech \citep{ren_fastspeech_2019} and FastSpeech 2 \citep{ren2021fastspeech2} used non-autoregressive architectures and included phoneme input for faster, more robust synthesis. Both types generally require 20-50 hours of well-annotated data per speaker.

More recent large-scale systems map input text directly to discrete audio tokens (neural codecs) for waveform decoding. These models achieve state-of-the-art naturalness but demand massive datasets. For instance, VALL-E \citep{wang2023neural} was trained on 60,000 hours of speech, and Step-Audio \citep{huang2025step} on 10.65 million hours. Recording such volumes from a single speaker would take decades to centuries. While multi-speaker datasets are feasible for high-resource languages, they are impossible for low-resource languages with few speakers. Additionally, the environmental cost of large-scale data collection and modelling is deeply concerning.

Multilingual and code-mixed TTS magnifies these challenges. Mel-spectrogram-based systems rely on language-specific phone inventories (e.g., ARPABET, Pinyin), requiring separate front-ends for each language. Neural codec systems avoid explicit phonemes but still need extensive per-language data. Both approaches lack a shared cross-linguistic representation, limiting data pooling, scalability, and phonetic controllability: correcting pronunciation or specifying phonetic detail is difficult due to the lack of such interface.

To address these bottlenecks in data efficiency, extensibility, and interpretability, we propose an input system grounded in phonological feature theory, aiming towards a more universal TTS framework.

\subsection{Phonological features}
The benefits of using phonological features rather than language-specific labels, or even the more universal IPA, are threefold:

\begin{enumerate}[label=(\arabic*)]
\item Compact representation: Phonological features provide a smaller, universal set across languages. While IPA uses over 60 symbols, our study proposes only 20 FUL features across languages.
\item Data efficiency: A unified feature set allows pooling data across languages, reducing training data requirements and mitigating sparsity, particularly for low-resource languages.
\item Unified acoustic modelling: Features enable multiple languages and speakers to be integrated into a single acoustic model, preserving timbre in code-mixed speech and simplifying deployment.
\end{enumerate}

Previous studies have explored phonological features in TTS \citep{gutkin_fonbund_2019, maniati_cross-lingual_2021, staib_phonological_2020, wells_cross-lingual_2021,louw2023cross}. \citet{staib_phonological_2020} used ten non-theory-driven features trained on English and Mexican Spanish corpora, successfully synthesising German speech not in the training set. \citet{gutkin_fonbund_2019} examined three feature sets -- PHOIBLE (37), PanPhon (23), PhonClassCounts (13) -- on a nine-language dataset spanning Indo-Aryan and Dravidian languages. Three out of five test languages showed improvements over a phonemic baseline; the two unsuccessful cases were attributed to suboptimal phone inventory design. \citet{maniati_cross-lingual_2021} used 23 articulatory-inspired IPA features on subsets of a six-language corpus (666.8 hours). The highest MOS (Mean Opinion Score, rated on a scale from 1 to 5, typically reflecting perceived naturalness) was 3.68 and occurred when the test and training languages were typologically similar; the lowest (1.70) occurred for typologically distant language pairs. \citet{wells_cross-lingual_2021} used 19 SPE features \citep{chomsky_sound_1968} plus five text-based features, training on English and/or German. With 14 hours of English and 15 minutes of German, models achieved naturalness comparable to models trained on four hours of German using phones or features. More recent work demonstrates that even smaller feature sets can support single-language synthesis \citep{taannander2024beyond} and model phonological conditions in disordered speech \citep{naslund2024hypernasality}.

\subsection{Featurally Underspecified Lexicon (FUL) Features}
\label{sec:FUL}
Distinctive features, first proposed by \citet{jakobsonPreliminariesSpeechAnalysis1952}, are designed to describe the phonological systems in the world's languages. Since then, distinctive feature theory has received sustained attention from phonologists, and various feature organisations have been proposed \citep[see][for detailed reviews]{Lahiri2010, lahiri_predicting_2018}. 

We chose the FUL model \citep{LahiriReetz2002, Lahiri2010, lahiri_predicting_2018} from among the many phonological feature theories and models. FUL defines a minimal inventory of distinctive features sufficient to represent phonemic contrasts across all languages. The framework has been investigated extensively in speech perception \citep[e.g.,][]{kotzorSymmetryAsymmetryEvidence2017}, the acquisition of phonological feature systems and developmental speech processing \citep[e.g.,][]{ghiniPlaceArticulationFirst2012,althausCoronalUnderspecificationEmerging2024}, loanword adaptation \citep[e.g.,][]{kennardNonesuchPhonemesLoanwords2020,lahiriPertinacityLoanwordsSame2019}, and, more recently, automatic speech recognition \citep{arora_phonological_2018}. To our knowledge, however, it has not previously been evaluated as a language-independent input representation for multilingual text-to-speech synthesis.

\begin{figure}[ht]

\centerline{\includegraphics[clip,trim=1cm 0cm 2cm 0,width=1.1\columnwidth]{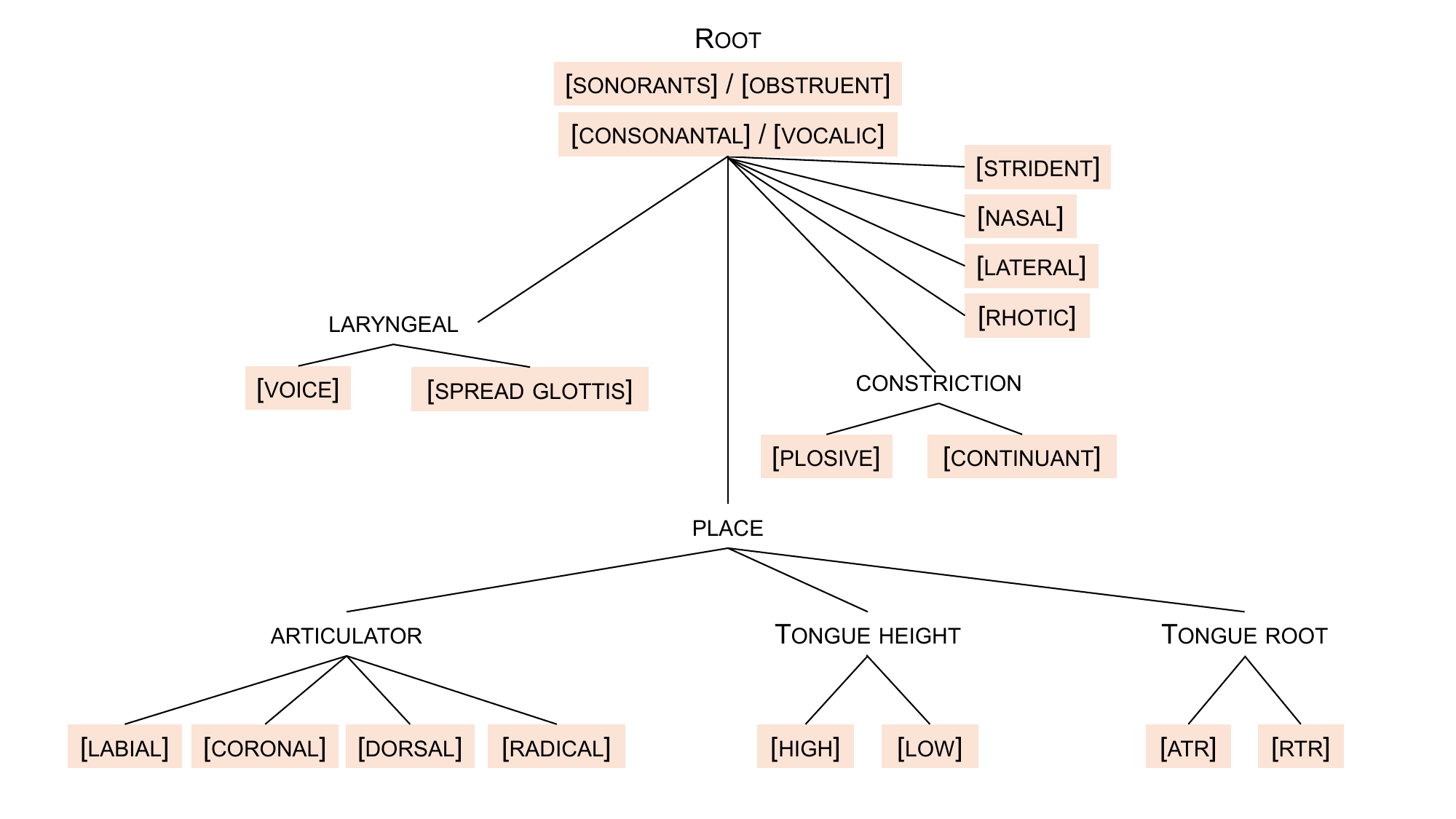}}

\caption{\label{fig:FIG1}Tree view of FUL features and nodes \citep[adapted from][]{Lahiri2010, lahiri_predicting_2018}. The shaded labels indicate the features in FUL.}
\end{figure}

Fig.~\ref{fig:FIG1} shows the FUL feature geometry, with all 20 features highlighted by shading. 
The FUL model has the following characteristics, which make it suitable for our purposes: 

\begin{itemize}[itemsep=12pt, leftmargin=2em]
    \item All features in the FUL model are monovalent, meaning they are either present or absent, without requiring binary oppositions such as [+voice] versus [-voice]. Within FUL, only contrastive, active features are specified in the lexicon, while unmarked properties are underspecified. For example, phonological processes such as final devoicing can be modeled by delinking the |voice| feature, rather than flipping a binary value from [+voice] to [-voice]. This both streamlines computation by reducing dimensionality, and aligns with articulatory models in which gestures are triggered by the presence of a feature rather than suppressed when absent.    

    \item  Consonants and vowels share \textsc{\lowercase{PLACE}} features. No extra feature is needed to describe vowels. For example, labial consonants and rounded vowels are [\textsc{\lowercase{LABIAL}}], coronal consonants and front vowels are [\textsc{\lowercase{CORONAL}}], and dorsal consonants and back vowels are [\textsc{\lowercase{DORSAL}}]. For consonants, the \textsc{\lowercase{TONGUE HEIGHT}} and \textsc{\lowercase{TONGUE ROOT}} features become relevant when there are contrasts within the same \textsc{\lowercase{ARTICULATOR}}. For instance, contrasts within coronal sounds, e.g., dentals, palatals vs. retroflexes, can be established by a combination of [\textsc{\lowercase{CORONAL}}] and \textsc{\lowercase{TONGUE HEIGHT}} features. The number of features is therefore also kept to a minimum. 
    
    \item The features in the geometry have acoustic correlates; e.g., the \textsc{\lowercase{TONGUE HEIGHT}} feature [\textsc{\lowercase{HIGH}}] is characterised by the concentration of more energy at higher frequencies and a low F1, and vice versa for the feature [\textsc{\lowercase{LOW}}]. This allows FUL features to account for all world languages.  
  
\end{itemize}


\subsection{FUL features for Mandarin and English}

\subsubsection{English}

The features for the phones in General American English in this study are shown in Table~\ref{tab:table1}. The parentheses indicate that the feature specification for the phone is optional as it is not required to establish contrast within the language, but it is specified in this study since our aim is to build a cross-linguistic system which can distinguish both English and Mandarin. In English, there are 38 phones, within which 24 are [\textsc{\lowercase{CONSONANTAL}}] and 14 are [\textsc{\lowercase{VOCALIC}}].

Consonants are further divided up by the \textsc{\lowercase{ROOT}} features, [\textsc{\lowercase{OBSTRUENT}}] and [\textsc{\lowercase{SONORANT}}], into obstruents /b, p, v, f, d, t, \textipa{\textdyoghlig}, \textteshlig, \textipa{\dh}, \texttheta, s, z, \textesh, \textyogh, \textscriptg, k, h/ and sonorants /m, n, \textipa{\ng}, w, j, l, \textturnr/ respectively. Obstruents contrast in voicing. Voiced obstruents /b, v, d, \textipa{\textdyoghlig}, \textipa{\dh}, z, \textyogh , \textscriptg/ are specified for [\textsc{\lowercase{VOICE}}] and voiceless obstruents /p, f, t, \textteshlig, \texttheta, s, \textesh, k, h/ are not specified for \textsc{\lowercase{LARYNGEAL}} features. The plosives, fricatives, and affricates are distinguished by the feature [\textsc{\lowercase{STRIDENT}}] and the \textsc{\lowercase{CONSTRICTION}} features [\textsc{\lowercase{PLOSIVE}}] and [\textsc{\lowercase{CONTINUANT}}]. Plosives /b, p, d, t, \textscriptg, k/ are [\textsc{\lowercase{PLOSIVE}}]. Fricatives /v, f, \textipa{\dh}, \texttheta, s, z, \textesh, \textyogh, h/ are [\textsc{\lowercase{continuant}}], with /s, z, \textesh, \textyogh/ also being [\textsc{\lowercase{STRIDENT}}]. Affricates /\textipa{\textdyoghlig}, \textteshlig/ are [\textsc{\lowercase{PLOSIVE}}, \textsc{\lowercase{STRIDENT}}]. All sonorant consonants are specified for [\textsc{\lowercase{VOICE}}]. /m, n, \textipa{\ng}/ are [\textsc{\lowercase{NASAL}}], /l/ is [\textsc{\lowercase{LATERAl}}] and /\textturnr/ is [\textsc{\lowercase{RHOTIC}}]. The main place of articulation for the consonants is represented by the \textsc{\lowercase{ARTICULATOR}} features under the \textsc{\lowercase{PLACE}} node. The coronal fricative phones /s, z, \textesh, \textyogh/ contrast in their specification for \textsc{\lowercase{TONGUE HEIGHT}} features -- /\textesh, \textyogh/ are [\textsc{\lowercase{HIGH}}] and /s, z/ are not specified for \textsc{\lowercase{TONGUE HEIGHT}}. 

Vowels share the features under the \textsc{\lowercase{PLACE}} node with consonants. The front vowels /i, \textsci, e, \textepsilon, \ae/ are [\textsc{\lowercase{CORONAL}}], rounded back vowels /u, \textupsilon, \textopeno, o/ are [\textsc{\lowercase{LABIAL}}, \textsc{\lowercase{DORSAL}}] and the unrounded back vowel /\textscripta/ is [\textsc{\lowercase{DORSAL}}]. The high vowels /u, \textupsilon, i, \textsci,/ are [\textsc{\lowercase{HIGH}}], low vowels /\ae,  \textscripta/ are [\textsc{\lowercase{LOW}}], and mid vowels /\textopeno, o, e, \textepsilon, \textschwa, \textrhookschwa, \textrhookrevepsilon, \textturnv/ are not specified for \textsc{\lowercase{TONGUE HEIGHT}}. General American English also makes use of the \textsc{\lowercase{TONGUE ROOT}} features to establish the tense /u, o, i, e, \textturnv/ [\textsc{\lowercase{ATR}}] and lax vowel /\textupsilon, \textopeno, \textsci, \textepsilon, \textscripta, \textschwa, \ae/ [\textsc{\lowercase{RTR}}] contrast. The vowels /e, o/ are often realised as the diphthongs [e\textsci, o\textupsilon ], but they are usually represented as tensed monophthongs in the lexicon. By contrast, the true diphthongs in General American English, /\textscripta \textsci, \textscripta \textupsilon, \textopeno \textsci /, were represented as vowel sequences (/\textipa{A+I}/, /\textipa{A+U}/ and /\textipa{O+I}/, respectively) in the feature mapping, with each vowel mapped to its own feature bundle rather than assigning a dedicated feature vector to the diphthong itself. General American English also contains two [\textsc{\lowercase{RHOTIC}}] vowels /\textrhookschwa, \textrhookrevepsilon/.

\begin{table}[ht]
\small
\renewcommand{\arraystretch}{1}
\caption{Features for English phones. Parentheses indicate optional feature specifications not required for contrast in English that are included to support cross-linguistic compatibility.}
\label{tab:table1}

\begin{tabular}{p{0.26\columnwidth}p{0.24\columnwidth}p{0.4\columnwidth}}

Classes & Features    & English phones \\
\hline
\multirow{6}{*}{\textsc{root}}

&\lbrack\textsc{sonorant}\rbrack    
& m\ w\ n\ j\ l\ \textturnr \ \textipa{\ng} \ u\ \textupsilon \ \textopeno \ o\ i\ \textsci \ e\ \textepsilon \ \ae \ \textscripta \ \textschwa \ \textrhookschwa \ \textrhookrevepsilon \ \textturnv\\

&\lbrack\textsc{obstruent}\rbrack   
& b\ p\ v\ f\ d\ t\ \textdyoghlig \ \textteshlig \ \textipa{\dh} \ \texttheta \ s\ z\ \textesh \ \textyogh \ \textscriptg\ k\ h\\

&\lbrack\textsc{consonantal}\rbrack 
& b\ p\ v\ f\ d\ t\ \textdyoghlig \ \textteshlig \ \textipa{\dh} \ \texttheta \ s\ z\ \textesh \ \textyogh \ \textscriptg\ k\ h\ m\ w\ n\ j\ l\ \textturnr \ \textipa{\ng}\\

&\lbrack\textsc{vocalic}\rbrack     
& u\ \textupsilon \ \textopeno \ o\ i\ \textsci \ e\ \textepsilon \ \ae \ \textscripta \ \textschwa \ \textrhookschwa \ \textrhookrevepsilon \ \textturnv\\

&\lbrack\textsc{strident}\rbrack       
&\textdyoghlig \ \textteshlig \ s\ z\ \textesh \ \textyogh \\

&\lbrack\textsc{nasal}\rbrack       
& m n \textipa{\ng}\\

&\lbrack\textsc{lateral}\rbrack     
& l\\

&\lbrack\textsc{rhotic}\rbrack      
& \textturnr \ \textrhookschwa \ \textrhookrevepsilon  \\

\cline{1-3}
\multirow{1}{*}{\textsc{laryngeal}}

&\lbrack\textsc{voice}\rbrack       
& b\ v\ d\ \textdyoghlig \ \textipa{\dh} \ z\ \textyogh \ \textscriptg\ m\ w\ n\ j\ l\ \textturnr \ \textipa{\ng} \ u\ \textupsilon \ \textopeno \ o\ i\ \textsci \ e\ \textepsilon \ \ae \ \textscripta \ \textschwa \ \textrhookschwa \ \textrhookrevepsilon \ \textturnv\\

&\lbrack\textsc{spread glottis}\rbrack 
&   \\

\cline{1-3}
\multirow{2}{*}{\textsc{constriction}}
&\lbrack\textsc{plosive}\rbrack     
& b\ p\ d\ t\ \textdyoghlig \ \textteshlig \ \textscriptg\ k\\

&\lbrack\textsc{continuant}\rbrack  
& v\ f\ \textipa{\dh} \ \texttheta \ s\ z\ \textesh \ \textyogh \ h\\

\cline{1-3}
\multirow{4}{*}{\textsc{articulator}}
&\lbrack\textsc{labial}\rbrack      
& b\ p\ v\ f\ m\ w\ u\ \textupsilon \ \textopeno \ o\\

&\lbrack\textsc{coronal}\rbrack     
& d\ t\ \textdyoghlig \ \textteshlig \ \textipa{\dh} \ \texttheta \ s\ z\ \textesh \ \textyogh \ n\ j\ l\ \textturnr \ i\ \textsci \ e\ \textepsilon \ \ae \\

&\lbrack\textsc{dorsal}\rbrack      
& \textscriptg\ k\ w\ \textipa{\ng} \ u\ \textupsilon \ \textopeno \ o\ \textscripta \\

&\lbrack\textsc{radical}\rbrack     
& h\\

\cline{1-3}
\multirow{2}{*}{\textsc{tongue} \textsc{height}}
&\lbrack\textsc{high}\rbrack        
& \textesh \ \textyogh \ u\ \textupsilon \ i\ \textsci \ (\textdyoghlig \ \textteshlig \ \textscriptg\ k\ w\ j)\\

&\lbrack\textsc{low}\rbrack         
& \ae \ \textscripta\\

\cline{1-3}
\multirow{2}{*}{\textsc{tongue root}}
&\lbrack\textsc{atr}\rbrack         
& (u\ o\ i\ e\ \textturnv )\\

&\lbrack\textsc{rtr}\rbrack         
& \textupsilon \ \textopeno \ \textsci \ \textepsilon \ \textscripta \ \textschwa \ (\ae )\\
                                         
\end{tabular}

\end{table}

\subsubsection{Mandarin}
Table~\ref{tab:table2} shows the features of the 37 Mandarin phones, following \citet{zhang_hanyu_2009}. Of these, 21 are [\textsc{\lowercase{CONSONANTAL}}] and 16 are [\textsc{\lowercase{VOCALIC}}]. The overall number of sounds, and the number of vowels and consonants, are similar to that of General American English. Different from English, Mandarin obstruents contrast in aspiration instead of voicing. The aspirated obstruents /p\textsuperscript{h}, t\textsuperscript{h}, ts\textsuperscript{h}, \textrtailt \textrtails \textsuperscript{h}, k\textsuperscript{h}/ are specified for [\textsc{\lowercase{SPREAD GLOTTIS}}] and the unaspirated ones [p, f, t, s, \textrtails , ts, \textrtailt \textrtails , k, x] are not specified for \textsc{\lowercase{LARYNGEAL}} features. The only voiced obstruent /\textrtailz/ is analysed as a syllabic approximant \citep[e.g., ][]{Duanmu2000e} or an apical vowel \citep[e.g., ][]{Lee-Kim2014} in other studies. This study follows \citet{Duanmu2000e} and treats it as /\textrtailz/ [\textsc{\lowercase{CONSONANTAL, OBSTRUENT, VOICE}}]. Mandarin has a three-way coronal contrast -- dental [s, ts, ts\textsuperscript{h}] vs. alveolo-palatal [\textctc , t\textctc , t\textctc \textsuperscript{h}] vs. retroflex [\textrtails , \textrtailt \textrtails , \textrtailt \textrtails \textsuperscript{h}]. Although the alveolo-palatals and the dentals are represented by different pinyin letters and are often treated as separate phonemes, the alveolo-palatals are restricted to high front vowel contexts and have also been analysed as allophones of their dental counterparts (for a short review, see \citealt{Hauser2023}). /s, ts, ts\textsuperscript{h}/ are realised as [\textctc , t\textctc , t\textctc \textsuperscript{h}] before the high front vowels /i, y/, and as [s, ts, ts\textsuperscript{h}] elsewhere. Therefore, in this study, we adopt an economical representation in line with FUL and use the same features to represent the alveolo-palatals and their corresponding dentals. The dental and retroflex phones are distinguished by the \textsc{\lowercase{TONGUE HEIGHT}} features. The retroflexes are marked as [\textsc{\lowercase{RTR}}] and the dentals are not specified for tongue height. The Mandarin sonorant consonants /m, n, \textipa{\ng}, w, j, l/ have the same features as those in English. Mandarin does not have any [\textsc{\lowercase{RHOTIC}}] consonant. Mandarin has a larger number of nasalised vowels and rhotic vowels than English. /u\textrhoticity, o\textrhoticity, \textepsilon\textrhoticity, a\textrhoticity, \textschwa\textrhoticity/ are [\textsc{\lowercase{RHOTIC}}] and /\~a\textrhoticity, \~\textschwa\textrhoticity, \~u\textrhoticity/ [\textsc{\lowercase{NASAL, RHOTIC}}]. Besides the back rounded vowels that also exist in English, Mandarin has a front rounded vowel /y/ [\textsc{\lowercase{LABIAL, CORONAL}}].

\begin{table}[ht]
\small
\renewcommand{\arraystretch}{1}
\caption{Features for Mandarin phones. Parentheses indicate optional feature specifications not required for contrast in Mandarin that are included to support cross-linguistic compatibility.}

\label{tab:table2}

\begin{tabular}{p{0.26\columnwidth}p{0.24\columnwidth}p{0.4\columnwidth}}
Classes & Features    & Mandarin phones \\
\hline

\multirow{8}{*}{\textsc{root}}

&\lbrack \textsc{sonorant}\rbrack
& m w n j l \textipa{\ng} i\ y\ u\ u\textrhoticity\ \~u\textrhoticity\ o\ o\textrhoticity\ e\ \textepsilon \ \textepsilon \textrhoticity\ a\ a\textrhoticity\ \~a\textrhoticity\ \textschwa \ \textschwa\textrhoticity \ \~\textschwa\textrhoticity\\

&\lbrack \textsc{obstruent}\rbrack       
& p\ p\textsuperscript{h}\ f\ t\ t\textsuperscript{h}\ s\ \textrtails \ \textrtailz \ ts\ ts\textsuperscript{h}\ \textrtailt \textrtails \ \textrtailt \textrtails \textsuperscript{h}\ k\ k\textsuperscript{h}\ x\\

&\lbrack \textsc{consonantal}\rbrack     
& p\ p\textsuperscript{h}\ f\ t\ t\textsuperscript{h}\ s\ \textrtails \ \textrtailz \ ts\ ts\textsuperscript{h}\ \textrtailt \textrtails \ \textrtailt \textrtails \textsuperscript{h}\ k\ k\textsuperscript{h}\ x\ m\ w\ n\ j\ l\ \textipa{\ng} \\

&\lbrack \textsc{vocalic}\rbrack        
& i\ y\ u\ u\textrhoticity\ \~u\textrhoticity\ o\ o\textrhoticity\ e\ \textepsilon \ \textepsilon \textrhoticity\ a\ a\textrhoticity\ \~a\textrhoticity\ \textschwa \ \textschwa\textrhoticity \ \~\textschwa\textrhoticity\\

&\lbrack \textsc{strident}\rbrack       & s\ \textrtails \ \textrtailz \ ts\ ts\textsuperscript{h}\ \textrtailt \textrtails \ \textrtailt \textrtails \textsuperscript{h}\\

&\lbrack \textsc{nasal}\rbrack          & m n \textipa{\ng} \~u\textrhoticity\ \~a\textrhoticity\  \~u\textrhoticity  \~\textschwa\textrhoticity\\

&\lbrack \textsc{lateral}\rbrack        & l\\

&\lbrack \textsc{rhotic}\rbrack         & u\textrhoticity\ \~u\textrhoticity\ o\textrhoticity \ \textepsilon\textrhoticity\ a\textrhoticity\ \~a\textrhoticity\ \textschwa\textrhoticity \ \~\textschwa\textrhoticity \\ 

\cline{1-3}
\multirow{2}{*}{\textsc{laryngeal}}

&\lbrack\textsc{voice}\rbrack          
& \textrtailz\ m w n j l \textipa{\ng} i\ y\ u\ u\textrhoticity\ \~u\textrhoticity\ o\ o\textrhoticity\ e\ \textepsilon \ \textepsilon \textrhoticity\ a\ a\textrhoticity\ \~a\textrhoticity\ \textschwa \ \textschwa\textrhoticity \ \~\textschwa\textrhoticity\\

&\lbrack\textsc{spread glottis}\rbrack & p\textsuperscript{h}\ t\textsuperscript{h}\ ts\textsuperscript{h}\ \textrtailt \textrtails \textsuperscript{h}\ k\textsuperscript{h}                                         \\

\cline{1-3}

\multirow{2}{*}{\textsc{constriction}}

&\lbrack\textsc{plosive}\rbrack        
& p\ p\textsuperscript{h}\ t\ t\textsuperscript{h}\ ts\ ts\textsuperscript{h}\ \textrtailt \textrtails \ \textrtailt \textrtails \textsuperscript{h}\ k\ k\textsuperscript{h}\\

&\lbrack\textsc{continuant}\rbrack     
& f\ s\ \textrtails \ \textrtailz \ x                                                  \\

\cline{1-3}
\multirow{3}{*}{\textsc{articulator}}

&\lbrack\textsc{labial}\rbrack         & p\ p\textsuperscript{h}\ f\ m\ w\ y\ u\ u\textrhoticity\ \~u\textrhoticity
  o\ o\textrhoticity                             \\

&\lbrack\textsc{coronal}\rbrack        & t\ t\textsuperscript{h}\ s\ \textrtails \ \textrtailz \ ts\ ts\textsuperscript{h}\ \textrtailt \textrtails \ \textrtailt \textrtails \textsuperscript{h}\ n\ j\ l\ i\ y\ e \textepsilon \ \textepsilon \textrhoticity \\

&\lbrack\textsc{dorsal}\rbrack         & k\ k\textsuperscript{h}\ x\ w\ \textipa{\ng} \ u\ u\textrhoticity\ \~u\textrhoticity o\ o\textrhoticity\\

&\lbrack\textsc{radical}\rbrack     
& \\

\cline{1-3}
\multirow{2}{*}{\textsc{tongue height}}

&\lbrack\textsc{high}\rbrack            & \textrtails \ \textrtailz \ \textrtailt \textrtails \ \textrtailt \textrtails \textsuperscript{h}\ i\ y\ u\ u\textrhoticity\ \~u\textrhoticity (k\ k\textsuperscript{h}\ w\ j)\\

&\lbrack\textsc{low}\rbrack             & a\ a\textrhoticity\ \~a\textrhoticity \\

\cline{1-3}
\multirow{2}{*}{\textsc{tongue root}}

&\lbrack\textsc{atr}\rbrack             & (i y u\ u\textrhoticity\ \~u\textrhoticity o\ o\textrhoticity e )\\

&\lbrack\textsc{rtr}\rbrack             & \textrtails \ \textrtailz \ \textrtailt \textrtails \ \textrtailt \textrtails \textsuperscript{h}\ \textepsilon \ \textepsilon \textrhoticity \ \textschwa\textrhoticity \ \~\textschwa\textrhoticity\\

\end{tabular}

\end{table}

\section{Methods}

\subsection{Phone conversion}
\label{sec:phone_conversion}

We first created a mapping \citep[available in][]{zhangPhonologicalFeatureMapping2021} between language-specific phone systems. For English, we used ARPABET and converted it to SAMPA \citep{wellsSAMPAComputerReadable1997}. For Mandarin, we used Pinyin and converted it to SAMPA-SC \citep{zhang_hanyu_2009}. Each SAMPA symbol was then mapped to a set of FUL features according to Tables~\ref{tab:table1} and~\ref{tab:table2}. The final dataframe included a column for the SAMPA symbols and one column for each of the 20 FUL features. When a feature is applicable to a phone, a value of “1” indicates its presence.

The conversion to SAMPA was optional. SAMPA was only used as an intermediate transcription system because it provides a flexible and language-independent interface. Many existing resources, such as pronunciation dictionaries, G2P resources, and annotations, use ARPABET for English or Pinyin for Mandarin, which can be easily converted into SAMPA. Unlike ARPABET, which is primarily designed for American English, and Pinyin, which is specific to Mandarin Chinese, SAMPA covers a much broader range of languages and phonetic distinctions. This cross-linguistic compatibility allows existing resources to be reused efficiently and makes it easier to extend the system to additional languages. For languages without an established SAMPA inventory, the binary feature vectors can be generated directly, making the SAMPA stage optional. 

\subsection{Model configuration}

\begin{figure}[ht]
  \centerline{\includegraphics[width=\columnwidth]{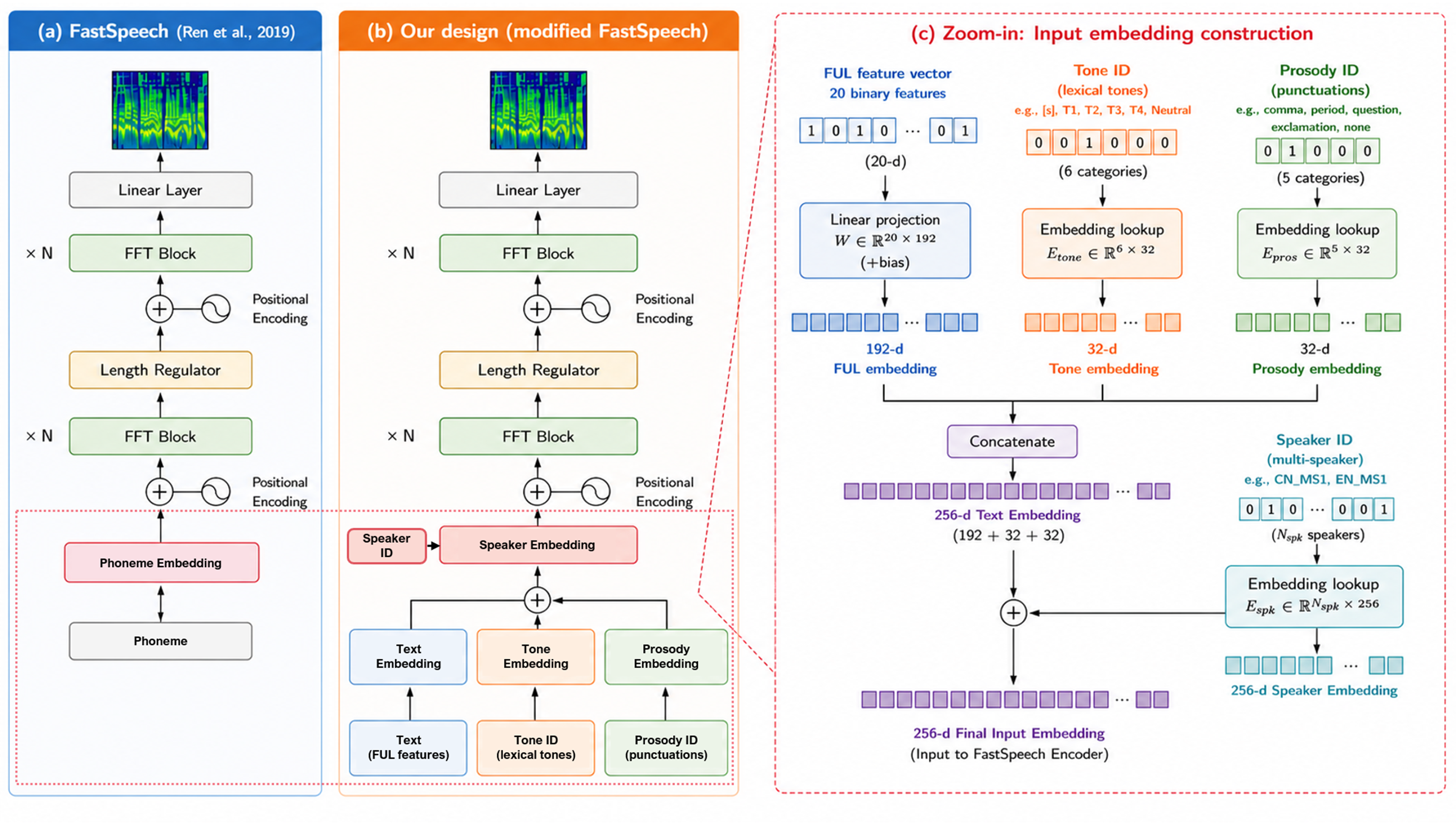}}
  \caption{\label{fig:FIG2} Model architecture and input embedding design. (a) Original FastSpeech architecture \citep{ren_fastspeech_2019}. (b) Modified FastSpeech architecture used in this study. The region enclosed by the red dotted box highlights the modified input embedding component. (c) Zoomed-in view of the input embedding used in this study.}
\end{figure}

We adopted a modified FastSpeech architecture \citep{ren_fastspeech_2019}, replacing the phoneme-only input layer with a combined text and speaker embedding, as shown in Figure~\ref{fig:FIG2}. The modified input components are highlighted by the red dotted box in panels (a) and (b), with a zoomed-in view of the proposed input representation shown in panel (c). The combined text input comprised three representations: (1) \textit{FUL features}: phonological features for segmental pronunciation, converted using the mapping introduced in Section~\ref{sec:phone_conversion}, see also \citet{zhangPhonologicalFeatureMapping2021}; (2) \textit{Tone}: lexical tone labels where applicable (five lexical tonal categories including four full tones and a neutral tone, plus any silences indicated by [sil]); and (3) \textit{Prosody}: prosodic features were derived from punctuations, including commas, periods, question marks, and exclamation marks. The phonological feature vectors were first projected to a 192-dimensional space via a linear layer. \replaced{The tone labels and prosodic labels were first converted into learnable 32-dimensional embedding vectors through embedding lookup tables. As these inputs comprise a small number of categorical labels, a 32-dimensional embedding was found to provide a lightweight yet sufficient representation while preserving the overall input dimensionality of the original FastSpeech architecture.}{The tone features and prosodic features were embedded into 32-dimensional vectors each.} The three embeddings were concatenated to form a 256-dimensional text embedding. For multi-speaker training, each speaker was assigned a unique speaker ID, which was mapped to a learnable \added{256-dimensional} speaker embedding. The speaker embedding was added to the text embedding before being passed to the encoder, enabling the model to control speaker identity during synthesis. \added{The overall embedding dimensionality was intentionally kept identical to that of the original FastSpeech architecture so that any performance differences could be attributed to the proposed input representation rather than changes in model capacity.} The rest of the FastSpeech architecture, including the encoder and decoder, remained unchanged. The number of FFT blocks on both the phoneme side and the mel-spectrogram side were set to four. Temporal alignment of the phones was obtained using the Kaldi ASR toolkit \citep{poveyKaldiSpeechRecognition2011}.

We trained our modified FastSpeech model on an NVIDIA GeForce RTX 2080 GPU with a batch size of 24. The training lasted for 200k steps until convergence. The optimiser and other hyperparameters were the same as those reported in \citet{ren_fastspeech_2019}. The model size was also comparable to the original FastSpeech architecture, as the main modification lies in the input representation rather than the core network structure. In our implementation, the model contained approximately 27 million parameters. During inference, the system achieves a real-time factor (RTF) of approximately 0.03 on the GPU (excluding the vocoder). The output mel-spectrograms were transformed into audio samples using a pretrained MelGAN vocoder \citep{kumarMelGANGenerativeAdversarial2019}.

\section{Experiments}
\added{The experiments are organised in two stages. Experiment~1 uses a small dataset and varies the amount of Mandarin training data while keeping the English data constant. Experiment~2 uses a larger, multi-speaker dataset to examine intelligibility across a broader set of voices and language conditions.} Audio demos of the outputs from the following two experiments are publicly available\footnote{\url{https://congzhang365.github.io/feature_tts/}}.

\subsection{Experiment 1: Small Dataset}

This experiment investigated whether phonological features could support intelligible speech synthesis with minimal but incremental training data. We evaluated model performance on Mandarin, English, and Mandarin-English code-mixed utterances. 

\subsubsection{Data}

The training data consisted of studio-recorded, high-quality audio originally acquired for commercial TTS development. \added{Professional voice actors were recruited and compensated by the company to record a large corpus of prompted sentences in a controlled studio environment. The recordings were available to the authors through research collaboration.} All recordings were sampled at 24 kHz with 16-bit resolution. Data from two speakers were used in this experiment: one in Mandarin and one in English.

The English dataset was recorded by a male native speaker of American English, styled as an AI assistant. This speaker is referred to as \texttt{EN\_MS1}, encoding language and gender. The total duration of this dataset was 2.62 hours.

The Mandarin dataset was recorded by a male native speaker of Mandarin, styled as a newsreader. This speaker is referred to as \texttt{CN\_MS1}. The full dataset comprised 20 hours of speech.

It is important to note that each speaker is monolingual in the training data: \texttt{EN\_MS1} appears only in English recordings, while \texttt{CN\_MS1} appears only in Mandarin. At inference time, however, each model is conditioned to synthesise both languages using either speaker voice. This results in cross-lingual synthesis conditions in which a given speaker is required to produce a language not observed for that speaker during training.

To assess the impact of training data size, we held the English data constant and varied the amount of Mandarin data across three conditions (labelled by the total hours of Mandarin data used):
\begin{itemize}
    \item \textbf{M0.5}: comprising 0.5 hours of Mandarin from \texttt{CN\_MS1} + 2.62 hours of English from \texttt{EN\_MS1}
    \item \textbf{M2}: comprising 2 hours of Mandarin from \texttt{CN\_MS1} + 2.62 hours of English from \texttt{EN\_MS1}
    \item \textbf{M8}: comprising 8 hours of Mandarin from \texttt{CN\_MS1} + 2.62 hours of English from \texttt{EN\_MS1}
\end{itemize}

This design allowed us to examine how increasing Mandarin input affected synthesis quality in both seen and unseen languages for each voice, as well as in code-mixed utterances.

\subsubsection{Evaluation}

Each model was capable of generating output using either speaker’s timbre. Accordingly, three models were trained: \texttt{M0.5}, \texttt{M2}, and \texttt{M8}. Each model was capable of synthesising speech using both \texttt{CN\_MS1} and \texttt{EN\_MS1} voices, yielding six evaluation conditions. We generated 78 test utterances \replaced{in total}{per model} for evaluation, comprising five Mandarin-only, five English-only, and three Mandarin-English code-mixed sentences\footnote{The list of test utterances can be found in the OSF repository for this project: \url{https://osf.io/c78nz/}}.

\replaced{The primary goal of this experiment was to assess whether phonological feature-based input could yield intelligible speech, even with limited training data. To this end, we collected subjective intelligibility ratings (henceforth referred to as \textit{intelligibility MOS} to differentiate from MOS on naturalness).}{The primary goal of this experiment was to assess whether phonological feature-based input could yield intelligible speech, even with limited training data. To this end, we conducted a subjective intelligibility rating, as the primary objective was to determine whether incorporating FUL features into the input representation, even with a limited dataset, could produce intelligible speech. Therefore, we collected subjective intelligibility ratings (henceforth referred to as \textit{intelligibility MOS} to differentiate from MOS on naturalness).} Listeners rated the model outputs on a 5-point scale (1 = least intelligible, 5 = most intelligible). The experiment was conducted online using Psytoolkit \citep{stoetPsyToolkitSoftwarePackage2010, stoetPsyToolkitNovelWebbased2017}. \added{All listeners participated voluntarily and provided informed consent.} They evaluated outputs from all models. Utterances from different models and conditions were presented in a fully randomised order for each participant. No textual transcriptions were provided; listeners rated intelligibility based solely on the audio. No qualitative feedback was collected from the participants. 

The English output was rated by 16 native English speakers, while the Mandarin and code-mixed outputs were rated by 19 native Mandarin speakers, all of whom were also proficient in English. The presentation order of all utterances was randomised for each participant to minimise potential order and repetition effects. The same group of listeners participated in both Experiment 1 and Experiment 2. No demographic information was collected from the participants, as the experiment was conducted anonymously online \added{through Psytoolkit} and no further analyses were planned.

\subsubsection{Results}
\label{sec:result1}

\begin{figure}[htb]
  \centering
  \includegraphics[width=1\linewidth]{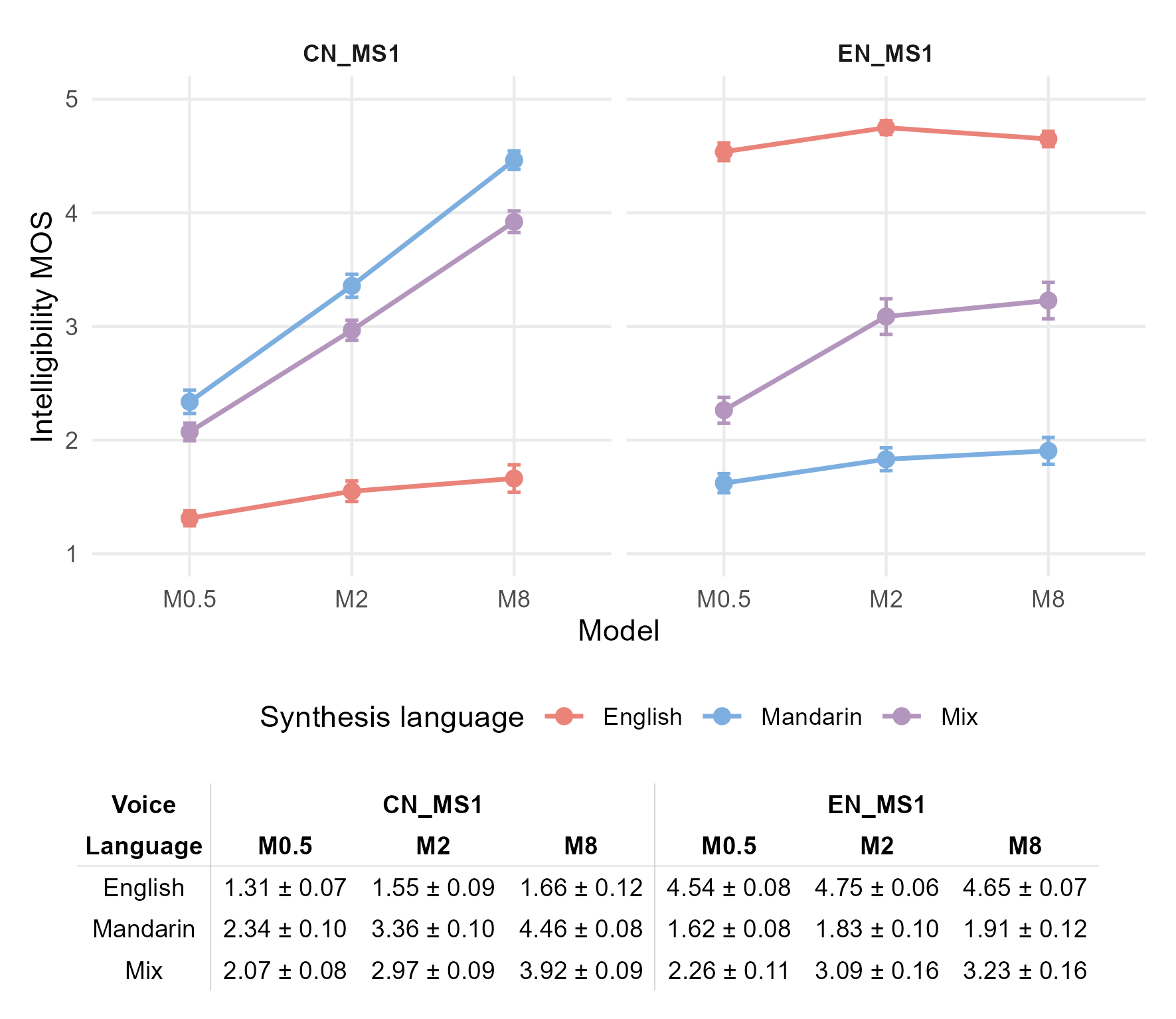}
  \caption{Mean intelligibility MOS for English, Mandarin, and Mandarin-English code-mixed speech synthesised using the Mandarin voice (\texttt{CN\_MS1}) and English voice (\texttt{EN\_MS1}). M0.5, M2, and M8 denote models trained with 0.5, 2, and 8 hours of Mandarin speech, respectively; all models additionally included 2.62 hours of English speech. Points represent mean MOS ratings and error bars indicate ±1 standard error.}
  \label{fig:exp1}
\end{figure}

We fitted cumulative link mixed models (CLMMs) to the intelligibility MOS ratings using the \texttt{clmm} function from the \texttt{ordinal} package \citep{ordinal} in R \citep[][version 4.4.2]{R-base}. Fixed effects included \texttt{Model} (\texttt{M0.5}, \texttt{M2}, \texttt{M8}), \texttt{Voice} (\texttt{CN\_MS1}, \texttt{EN\_MS1}), and \texttt{SynthesisLanguage} (English, Mandarin, Mix), along with their interactions. \deleted{All predictors were within-participant variables.} \texttt{Model} was Helmert-coded and the other fixed effects were sum-coded. Random intercepts for \texttt{participant} and \texttt{stimuli} were included, along with by-participant random slopes for all predictors in the full model. The final successfully converged model was specified with random slopes for \texttt{Model} and \texttt{Voice} by \texttt{participant}.\footnote{More detailed analyses and results for both experiments can be found in the OSF repository for this project: \url{https://osf.io/c78nz/}}.

The model revealed a significant main effect of \texttt{Model}, with intelligibility increasing as more training data was included (both contrasts $p < .001$). A significant main effect of \texttt{Voice} was also observed ($p < .001$), indicating overall differences between the two speakers. Crucially, there was a strong interaction between \texttt{Voice} and \texttt{SynthesisLanguage} ($p < .001$). In addition, \texttt{Model} interacted with both \texttt{Voice} and \texttt{SynthesisLanguage}, indicating that the effect of increased training data varied across conditions. No significant three-way interaction was found.

To further examine these effects, post hoc comparisons were conducted using the \texttt{emmeans} package \citep{emmeans}. 

\added{In addition to the quantitative analyses reported below, we include a small number of qualitative observations based on expert listening. Following \citet{sanchez2025}, we use expert listening to refer to subjective evaluation conducted by the authors. Such observations are commonly used in TTS development to complement formal evaluations by highlighting perceptually salient phenomena that may not be fully reflected in quantitative measures. In the present study, these observations were made primarily by the first author, a trained phonetician who is Chinese-English bilingual, has lived in the UK for over 15 years, and uses English as her primary working language. The observations were subsequently discussed with and confirmed by the remaining co-authors, who are also Chinese-English bilinguals. These observations are intended to provide qualitative context and are not part of the formal evaluation.}

\textbf{English utterances}: Given the availability of 2.62 hours of native English training data, the \texttt{EN\_MS1} voice produced English utterances with consistently high intelligibility MOS across all three models (M0.5: 4.54, M2: 4.75, M8: 4.65, in Figure~\ref{fig:exp1}). However, the intelligibility MOS did not show a significant increase with more \added{Mandarin} training data, likely due to a ceiling effect.

In contrast, \texttt{CN\_MS1} produced English output with consistently low intelligibility ratings (MOS < 1.7), with no significant differences between conditions ($p > .05$). English was an unseen language for \texttt{CN\_MS1}, and increasing the amount of Mandarin training data from 0.5 to 8 hours did not result in a significant improvement in English intelligibility. This suggests limited cross-lingual generalisation from Mandarin to English phonological features within the current training data range.

\textbf{Mandarin utterances}: For the voice with only Mandarin training data (\texttt{CN\_MS1}), increasing training data in the seen language significantly improved intelligibility for the synthesised Mandarin utterances, with all pairwise comparisons between model conditions reaching significance (all $p < .01$). The mean intelligibility MOS increased from 2.3 in \texttt{M0.5} to 3.4 in \texttt{M2}, and further to 4.5 in \texttt{M8}, as shown in Figure~\ref{fig:exp1}. Additional qualitative observations \added{from the authors} are as follows. In \texttt{M0.5}, \texttt{CN\_MS1} produced only partially intelligible Mandarin utterances, with some words identifiable but others unclear. Both \texttt{M0.5} and \texttt{M2} exhibited issues with audio quality, lexical tone accuracy, and intonation. In \texttt{M8}, \texttt{CN\_MS1} produced more intelligible utterances with improved lexical tone accuracy, although some minor audio quality issues remained.

For \texttt{EN\_MS1}, increasing the unseen language input did not result in significant improvements in intelligibility for Mandarin outputs (all $p > .05$) and the intelligibility MOS remained very low across all models: all intelligibility MOS scores were below 2.0. However, \replaced{the authors also surmised that}{it was also observed that}, if a listener knew the text of the utterance beforehand, some words in the \texttt{M8} condition could be identified, whereas this was less evident in \texttt{M0.5}.

\textbf{Mandarin-English code-mixed utterances}: Similar to the results for Mandarin utterances, \texttt{CN\_MS1} showed significant improvement in intelligibility MOS as the training data increased (all $p < .01$). The increase in intelligibility MOS for code-mixed utterances (from 2.07 to 3.92) closely paralleled that observed for Mandarin utterances \added{from 2.34 to 4.46, (no statistical difference between the Mandarin and the Mix conditions for \texttt{CN\_MS1}, p = 0.74)}, as shown in Figure~\ref{fig:exp1}, while improvements for English were comparatively small (from 1.31 to 1.66). This suggests that the improvement in code-mixed utterances may be largely driven by the Mandarin component with the increase in Mandarin training data, and that the presence of English segments did not appear to substantially limit overall performance under the present conditions.

For the \texttt{EN\_MS1} voice, the intelligibility MOS for code-mixed utterances was \added{numerically} higher than that for Mandarin-only utterances in this voice across all models. However, the differences between models did not reach statistical significance (all $p > .05$). \replaced{The authors also observed that, in M0.5, both \texttt{EN\_MS1} (MOS = 2.26) and \texttt{CN\_MS1} (MOS = 2.07) produced code-mixed utterances with low intelligibility.}{We also observed the following in the outputs: in \texttt{M0.5}, both \texttt{EN\_MS1} and \texttt{CN\_MS1} produced code-mixed utterances with low intelligibility.} In \texttt{M8}, \texttt{CN\_MS1} showed clear improvements in intelligibility, reaching a mean MOS of 3.92. For \texttt{EN\_MS1}, although the authors observed that the outputs became more fluent and segmentally clearer as the model was trained with increasing amounts of Mandarin data, this did not correspond to a statistically significant improvement in intelligibility MOS, which reached 3.23 in \texttt{M8}. This may be related to the absence of lexical tone specification on Mandarin syllables, which could reduce intelligibility despite improved segmental realisation.

\subsubsection{Interim Summary}
\added{Experiment 1 showed that increasing the amount of Mandarin training data substantially improved the intelligibility of Mandarin speech synthesised using the Mandarin speaker (\texttt{CN\_MS1}). Similar improvements were observed for Mandarin-English code-mixed speech, suggesting that intelligibility gains transferred effectively to code-mixed utterances. In contrast, intelligibility improvements for English speech synthesised using the Mandarin speaker were comparatively small, indicating that additional Mandarin training alone did not substantially benefit cross-lingual English synthesis. For the English speaker (\texttt{EN\_MS1}), increasing the amount of Mandarin training produced little improvement for either Mandarin or code-mixed speech, demonstrating limited cross-speaker transfer. Together, these findings suggest that phonological features learned from increased Mandarin training primarily benefit speech containing substantial Mandarin content, while cross-lingual generalisation remains more limited with such small datasets.}

\added{As Experiment~1 did not provide sufficient evidence of learning for an unseen language due to the relatively small increases in training data, Experiment~2  used a larger multi-speaker dataset comprising both English and Mandarin speech to further investigate whether phonological feature-based input could support intelligible speech synthesis across a broader range of speakers and language conditions.}

\subsection{Experiment 2: Larger Multi-Speaker Dataset}

\deleted{As Experiment~1 did not provide sufficient evidence of learning for an unseen language due to the relatively small increases in training data, we used a larger multi-speaker dataset comprising both English and Mandarin speech to further investigated whether phonological feature-based input could support intelligible speech synthesis across a broader range of speakers and language conditions.}

Subjective intelligibility ratings were collected employing the same listeners, test utterances, and procedures as in Experiment~1, with listeners also rating utterances on a 5-point intelligibility MOS scale. \added{Experiments~1 and~2 were administered through separate experiment links, allowing participants to complete them in separate sessions.}

Because the output in this experiment was more intelligible, automatic alignment was feasible. We therefore complemented the subjective ratings with an objective evaluation using the Charsiu textless phonetic alignment tool \citep{zhuPhonetoAudioAlignmentText2022}. The Phone Error Rate (PER) was calculated as follows:

\[
\text{PER} = \frac{S + I + D}{N}
\]

where \(S\) is the number of phoneme substitutions, \(I\) is the number of insertions, \(D\) is the number of deletions, and \(N\) is the total number of phonemes in the reference transcription. This metric provides a quantitative estimate of phoneme-level accuracy in the synthesised output.

\subsubsection{Data}

This experiment used a 100-hour multi-speaker dataset acquired for commercial TTS development. \added{Similar to Experiment 1, the data used here were also recorded by professional voice actors in a controlled studio environment, using prompted sentences.} The dataset included recordings from 12 speakers: nine native Mandarin speakers and three native speakers of General American English. The English data totalled 9 hours, while the Mandarin data comprised 76 hours of native Mandarin speech, 10 hours of Mandarin-English code-mixed speech, and 5 hours of second-language English speech produced by Mandarin speakers.

\added{The following three voices were selected to generate the test sentences, representing complementary training scenarios, including one cross-lingual speaker and two monolingual speakers, thereby enabling evaluation of both multilingual and zero-shot cross-lingual synthesis.}

\begin{itemize}
    \item \texttt{CN\_FS1}: A female native Mandarin speaker with 15 hours of cross-lingual data, including 10 hours of Mandarin, 2 hours of English, and 3 hours of Mandarin-English code-mixed speech. The recordings were styled for AI assistant applications.
    \item \texttt{CN\_MS2}: A male native Mandarin speaker with 9 hours of Mandarin speech, recorded in the style of public announcements and advertisements. This dataset does not contain any English data, so English is an unseen language for this voice.
    \item \texttt{EN\_MS1}: A male native speaker of American English with 2.62 hours of English speech, styled for AI assistant applications. This dataset does not contain any Mandarin data, so Mandarin is an unseen language for this voice. This speaker and dataset were also used in Experiment~1.
\end{itemize}

\subsubsection{Results}
\begin{figure}[htb]
  \centering
  \includegraphics[width=1\linewidth]{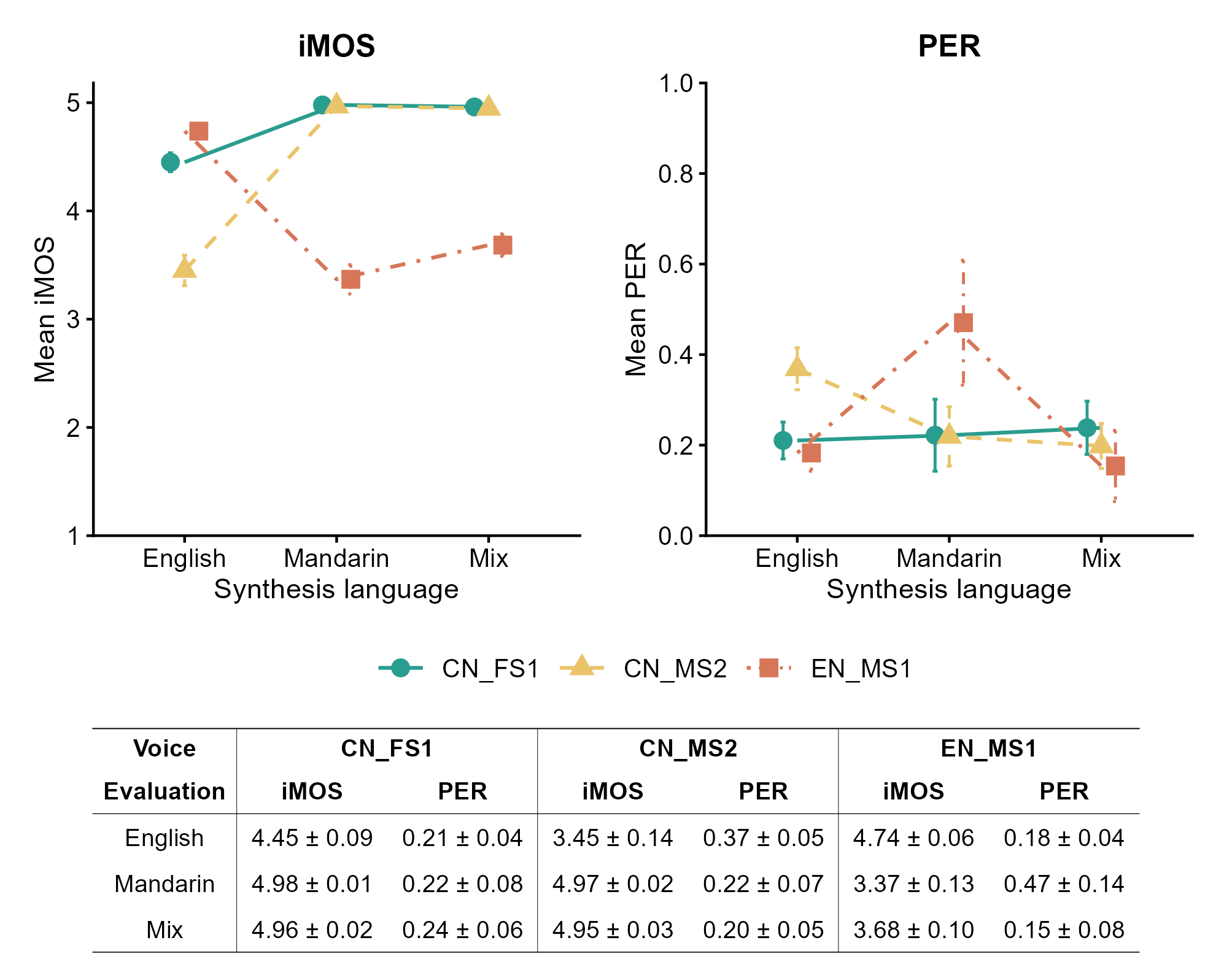}
  \caption{Mean intelligibility MOS (iMOS) and phoneme error rate (PER) for English, Mandarin, and Mandarin-English code-mixed speech synthesised using \texttt{CN\_FS1}, \texttt{CN\_MS2}, and \texttt{EN\_MS1}. Points represent mean values and error bars indicate ±1 standard error. Lines connect values from the same voice across synthesis languages; colours, point shapes, and line types distinguish the three voices.}
  \label{fig:exp2}
\end{figure}

Figure~\ref{fig:exp2} presents both subjective intelligibility MOS and objective phoneme error rates (PER) for the three selected voices. Trained on a larger, multilingual, multi-speaker 100-hour corpus, the models generally achieved higher intelligibility MOS scores than in Experiment~1.

\paragraph{Intelligibility MOS}
Following the analysis procedures used in Experiment~1, we fitted cumulative link mixed models (CLMMs) to the intelligibility MOS ratings for models in Experiment~2. Fixed effects included \texttt{Voice} (\texttt{CN\_FS1}, \texttt{CN\_MS2}, \texttt{EN\_MS1}) and \texttt{SynthesisLanguage} (English, Mandarin, Mix), along with their interaction. All predictors \deleted{were within-participant variables and}were sum-coded. Random intercepts for \texttt{Participant} and \texttt{Stimuli} were included, and by-\texttt{Participant} random slopes were specified for both \texttt{Voice} and \texttt{SynthesisLanguage} in the full model. The full model was retained as the final model. The model revealed differences between voices, although not all contrasts were statistically significant. Significant effects of SynthesisLanguage were observed ($p < .05$ for both contrasts). More importantly, there was a significant interaction between Voice and SynthesisLanguage ($p < .001$). To further examine these effects, post hoc comparisons were conducted using \texttt{emmeans}.

\textbf{English utterances}: \texttt{EN\_MS1} achieved higher intelligibility ratings than \texttt{CN\_MS2} ($p < .001$), and \texttt{CN\_FS1} also outperformed \texttt{CN\_MS2} ($p = .03$). The difference between \texttt{EN\_MS1} and \texttt{CN\_FS1} did not reach statistical significance. Consistent with this pattern, \texttt{CN\_FS1}, which included some English training data, produced English utterances with relatively high intelligibility ratings (MOS = 4.45), although slightly lower than those of \texttt{EN\_MS1} (MOS = 4.74). In contrast, \texttt{CN\_MS2}, which did not contain any English data, \replaced{received}{achieved} lower intelligibility ratings for English synthesis (MOS = 2.45) than \texttt{CN\_FS1} (MOS = 4.45) and \texttt{EN\_MS1} (MOS = 4.74). However, compared with Experiment~1, \texttt{CN\_MS2} showed a substantial improvement in English intelligibility, with the mean MOS increasing from 1.66 to 3.45. \added{As \texttt{CN\_MS2} is a different male speaker from \texttt{CN\_MS1}, this comparison is descriptive only and no direct statistical comparison between the two speakers can be made.} This suggests that, \replaced{for the model with the larger training dataset, the unseen language's intelligibility has also improved}{in the model with larger training dataset, the unseen language have also been improved in intelligibility}, and that the larger multi-speaker training data have facilitated some degree of cross-linguistic generalisation, allowing \texttt{CN\_MS2} to produce English utterances with non-trivial intelligibility despite the absence of direct English training data.

\begin{figure}[htb]
  \centering
  \includegraphics[width=0.8\linewidth]{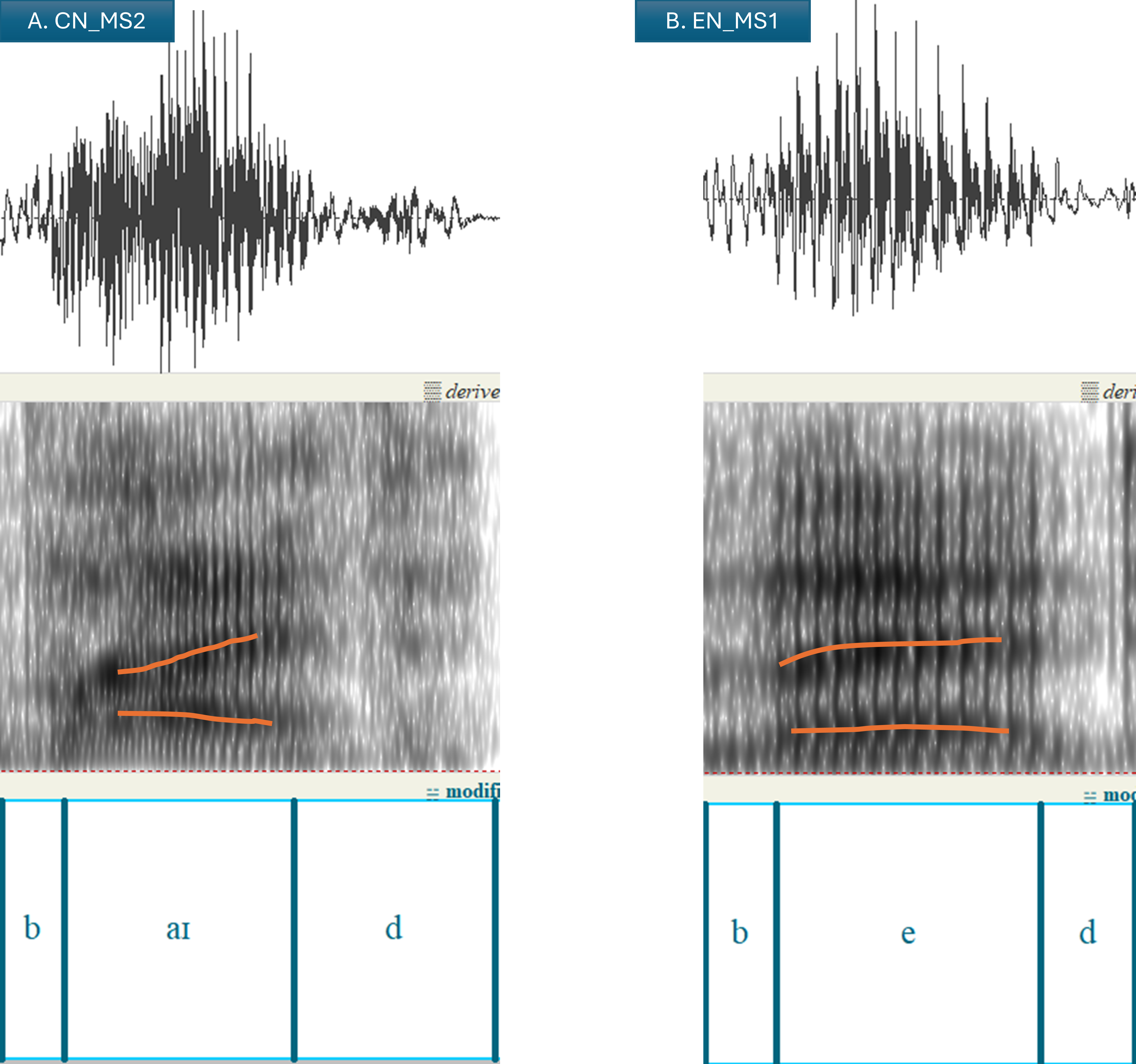}
  \caption{Example waveforms and spectrograms, with overlaid F1 and F2 formant tracks, for the English word `bed', synthesised with \texttt{CN\_MS2} and \texttt{EN\_MS1} voices in Experiment~2.}
  \label{fig:exp2a}
\end{figure}

Through examining the test utterances, we also observed that the English outputs from both Mandarin voices exhibited characteristics consistent with Mandarin-accented speech. For instance, Figure~\ref{fig:exp2a} compares the same word generated by a Mandarin voice (\texttt{CN\_MS2}) and the English voice (\texttt{EN\_MS1}), with F1 and F2 tracking highlighted. The Mandarin voice produced a diphthong [a\textipa{I}] for the vowel in `bed', as indicated by the gliding formants, whereas the English voice produced a more target-like monophthong with stable formants. This pattern is consistent with commonly reported substitutions in Mandarin-accented English, where a close-mid front vowel is not present in the Mandarin vowel inventory.

\textbf{Mandarin utterances}: The two Mandarin-trained voices, \texttt{CN\_FS1} and \texttt{CN\_MS2}, achieved significantly higher intelligibility ratings than the English voice (\texttt{EN\_MS1}) ($p < .001$), with no significant difference between the two Mandarin voices ($p > .9$). This pattern is consistent with the distribution of training data, as the Mandarin voices were trained on substantially more Mandarin speech, whereas the English voice had no direct exposure to Mandarin. \added{However, it is worth noting that, compared with Experiment~1, \texttt{EN\_MS1} showed substantially higher intelligibility for Mandarin utterances, with the mean MOS increasing from 1.62-1.91 to 3.37. This improvement suggests the presence of at least partial cross-language generalisation.}

\begin{figure}[htb]
    \centering
    \includegraphics[width=\linewidth]{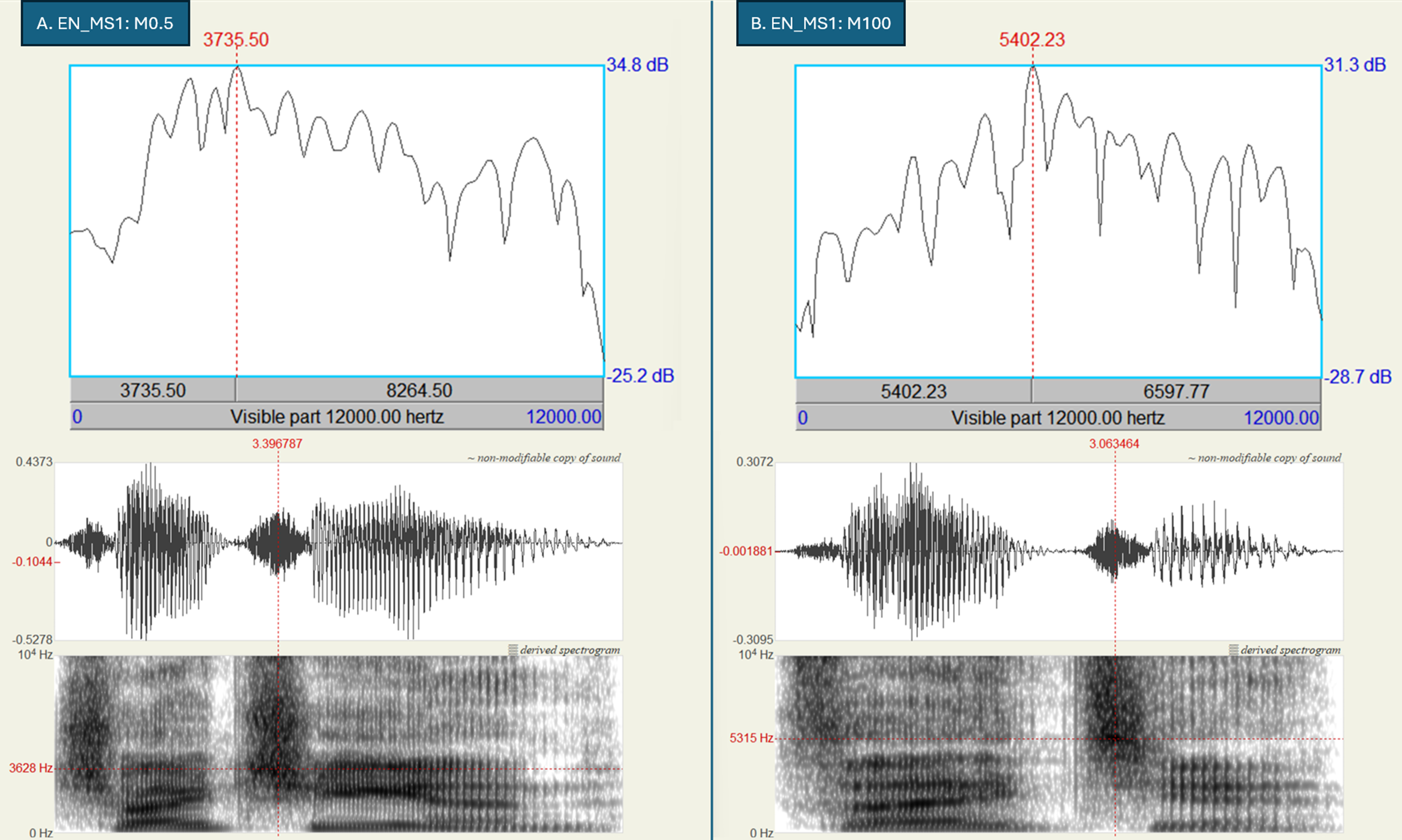}
    \caption{Spectral slice (top), waveform and spectrograms (bottom) for \textit{zàijiàn} (\textipa{/tsaI tCjEn/}, `goodbye'), synthesised with \texttt{EN\_MS1} voice in \texttt{M0.5} (Experiment~1) and \texttt{M100} (Experiment~2). }
    \label{fig:exp2b}
  \end{figure}

\replaced{As in Experiment 1, the following observations are based on expert listening (see Section~\ref{sec:result1}). As the primary focus of the present study is perceptual evaluation, the accompanying acoustic analyses are included only to illustrate selected perceptually salient observations rather than to provide a systematic acoustic analysis of all synthesised stimuli.}{Additional observations are as follows.} One notable pattern is that \texttt{EN\_MS1} was able to learn Mandarin segments through phonological features despite having no direct Mandarin data in its training set. Figure~\ref{fig:exp2b} compares the same word \textit{zàijiàn} (\textipa{/tsaI tCjEn/}, `goodbye') synthesised using \texttt{EN\_MS1} in the low-data condition of Experiment~1 (M0.5) and in the present 100-hour model. The spectral peak of the fricative portion of the affricate was 3735.50 Hz in the M0.5 model, and 5402.23 Hz in the M100 model. These values are consistent with the English postalveolar affricate \textipa{/\textdyoghlig/} \citep{maniwa2009} and Mandarin alveolo-palatal affricates \textipa{/t\textctc/} \citep{liAcousticAnalysisMandarin2015}, respectively. This suggests that with increasing training data, the model could learn to produce the Mandarin affricate with a more target-like place of articulation, shifting from a more English-like postalveolar articulation to a more target-like alveolo-palatal articulation. This indicates that the model can learn phonological features from different languages and apply them across languages, even for an unseen language.

\begin{figure}[htb]
  \centering
  \includegraphics[width=0.8\linewidth]{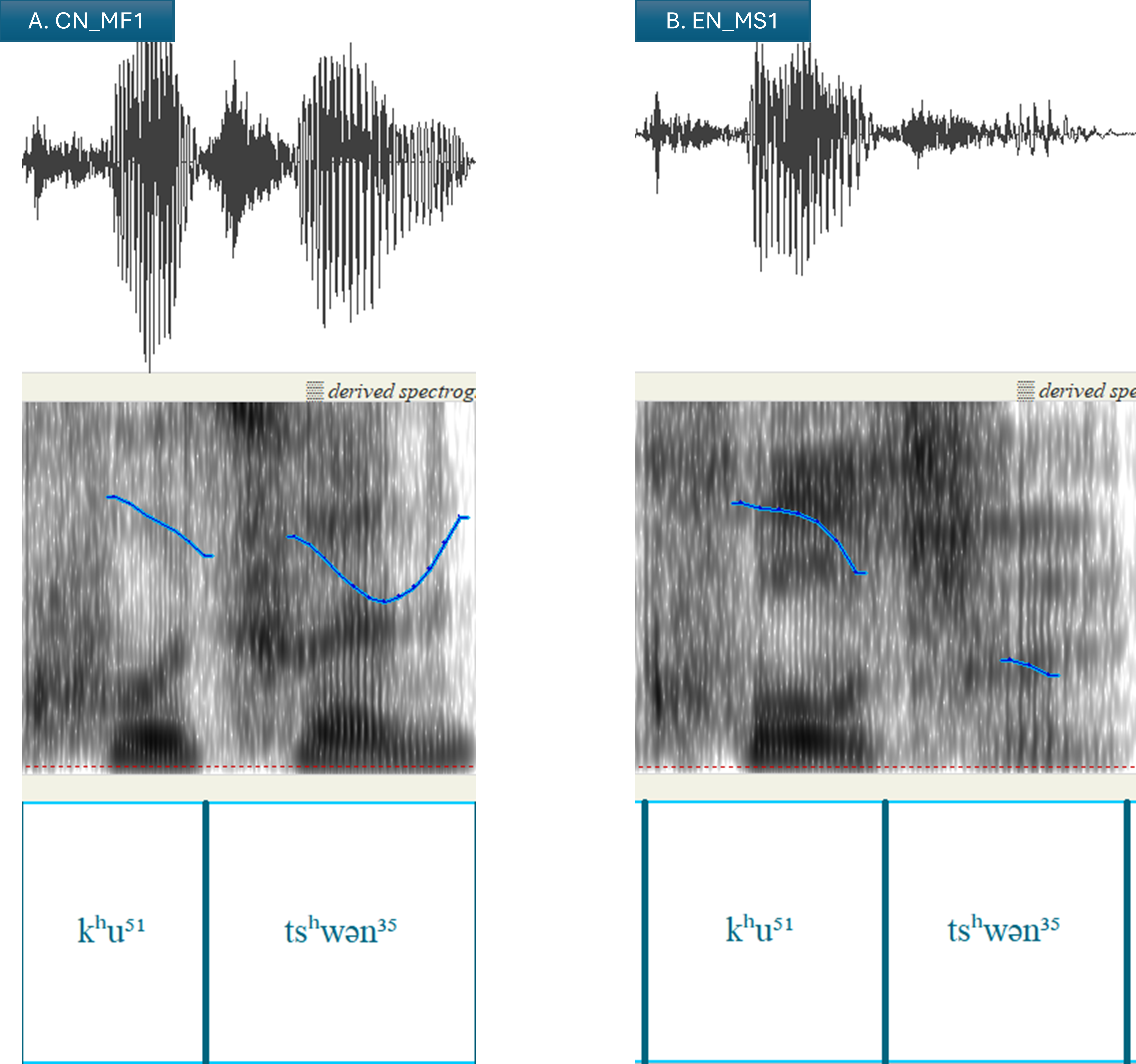}
  \caption{Example waveforms and spectrograms, with F0 tracks, for the word \textit{kùcún} (`stock') in Mandarin, synthesised with \texttt{CN\_FS1} and \texttt{EN\_MS1} voices in Experiment~2.}
  \label{fig:exp2c}
\end{figure}

  However, we also note that improvements in segmental realisation did not extend to tonal accuracy. As tonal information was only included in the Mandarin data using language-specific labels rather than feature-based representations, \texttt{EN\_MS1} was not able to reliably produce Mandarin lexical tones. This highlights the importance of developing feature-based prosodic representations to support cross-lingual generalisation for suprasegmental features such as tone. Figure~\ref{fig:exp2c} illustrates this with the word \textit{kùcún} (\textipa{/k\textsuperscript{h}u ts\textsuperscript{h}w\textipa{@}n/}, `stock'), comparing outputs from the Mandarin voice (\texttt{CN\_FS1}) and the English voice (\texttt{EN\_MS1}). The Mandarin voice produced a clear rising tone on the second syllable, whereas the English voice produced a short low f0 target, resembling a neutral tone rather than the intended rising tone. This lack of tonal specification is likely to have contributed to the lower intelligibility ratings for Mandarin utterances produced by the English voice, despite improvements in segmental accuracy.

\textbf{Mandarin-English code-mixed utterances}: Both \texttt{CN\_FS1} and \texttt{CN\_MS2} achieved significantly higher intelligibility ratings than \texttt{EN\_MS1} (both $p < .001$), with no significant difference between the two Mandarin speakers ($p > .9$). The mean intelligibility ratings for the two Mandarin voices were comparable and relatively high \added{(\texttt{CN\_FS1}: 4.96; \texttt{CN\_MS2}: 4.95)}. While this pattern is consistent with \texttt{CN\_FS1}'s exposure to code-mixed training data, it is noteworthy that \texttt{CN\_MS2} also achieved similarly high ratings despite the absence of both code-mixed and English training data. This again suggests that phonological feature-based representations can support cross-lingual generalisation, providing a shared representational space that facilitates the transfer of phonetic detail across languages.

\paragraph{Phone Error Rate (PER)}
We fitted a linear mixed-effects model to the PER scores using the \texttt{lme4} package \citep{lme4}, with \texttt{Voice} and \texttt{SynthesisLanguage} as fixed effects and \texttt{stimuli} as random intercept. Post hoc comparisons were again conducted using the \texttt{emmeans} package.

For English utterances, although \texttt{EN\_MS1} achieved the lowest mean PER and \texttt{CN\_MS2} the highest, the differences between voices did not reach statistical significance (all $p > .05$). This suggests that while there may be some variation in phonetic accuracy across voices for English synthesis, these differences are not robust enough to be statistically significant in this sample.

For Mandarin utterances, both Mandarin voices (\texttt{CN\_FS1} and \texttt{CN\_MS2}) achieved significantly lower PER scores than the English speaker (\texttt{EN\_MS1}) (both $p < .05$), with no significant difference between the two Mandarin speakers ($p > .9$). This pattern is consistent with the intelligibility MOS results and reflects the distribution of training data across voices. 

For the code-mixed utterances, again no significant differences were observed between voices (all $p > .05$), indicating comparable phonetic accuracy for the segments across speakers in this condition.

The PER results broadly align with the intelligibility MOS findings for Mandarin utterances, where both measures show that the Mandarin speakers outperform the English speaker in synthesising Mandarin utterances. However, the two measures diverge in the English condition and the code-mixed condition: while MOS results indicate some differences across voices, no significant differences were observed in PER. This suggests that while the phonetic accuracy of the synthesised segments may be comparable across voices, the perceived intelligibility can still differ, potentially due to factors such as prosody, audio quality, or other suprasegmental features that are not captured by PER.

\section{Discussion}
This study set out with three interrelated goals: 1) to evaluate the feasibility of applying phonological features in multilingual TTS, 2) to explore whether TTS can provide any empirical support for FUL theory, and 3) to present a mapping between widely used phone symbols and phonological features. \added{The present study is limited to a proof-of-concept evaluation using two languages. Further work is required to assess the generalisability of this approach across a wider range of languages, including genuinely low-resource settings, and to more fully evaluate its theoretical implications.}

\subsection{Feasibility of feature-based TTS}
With respect to the first aim, the results suggest that FUL-based input representations can support intelligible speech synthesis across both seen and unseen language conditions. In Experiment~1, even with relatively small datasets, models using FUL features as input were able to produce speech with moderate intelligibility in seen languages, i.e. the languages present in the training data for each voice. Experiment~2 further showed that increasing the amount of training data enabled the models to produce even unseen languages. Although performance in unseen languages remained lower than in seen languages, the intelligibility MOS scores indicated that the outputs were generally intelligible, even with only 100 hours of training data, \replaced{which is several orders of magnitude smaller than the tens of thousands to millions of hours of speech typically used to train contemporary state-of-the-art TTS models (see Section~\ref{sec:tts_modelling}).}{which is still a small number in TTS literature.} This study was not designed to compete with state-of-the-art TTS models; rather, it demonstrates that phonological features provide a theory-driven, data-efficient alternative for building multilingual TTS systems in low-resource scenarios. Feature-based representations also enable a level of controllability similar to that of earlier parametric TTS systems, as phonological features can be explicitly specified and manipulated. In contrast, end-to-end models typically rely on learned representations, making such control less direct and less interpretable.

Feature-based models provide a structured representation that may facilitate cross-lingual generalisation. This raises the possibility that such representations could support multilingual TTS in low-resource settings, where features learned from well-resourced languages transfer to under-represented ones. In principle, a relatively rare segment, such as a voiceless nasal, could be synthesised by recombining independently learned features from different languages. For instance, the absence of [voice] may be learned from voiceless obstruents, while the presence of [nasal] may be learned from common nasals such as an alveolar nasal /n/. Because these features are widely attested across languages, the model can generalise to their novel combination, even if the specific segment is absent from the training data. By contrast, phone-based input systems treat each segment as an atomic unit, requiring direct exposure to the target phone during training in order to synthesise it reliably. As a result, they may struggle to generate unseen or typologically rare segments in low-resource language varieties.

\subsection{Theoretical implications for FUL}

With respect to the second aim, the present study provides a new type of evidence for evaluating the FUL framework. In light of previous work on FUL as introduced in Section~\ref{sec:FUL}, the present findings extend the empirical scope of the framework by demonstrating that FUL features can serve as a viable representation for multilingual speech synthesis. The results show that FUL-based representations can support the generation of intelligible speech and a degree of cross-lingual transfer within a TTS system. Although a TTS system is inherently different from human speech production, the fact that FUL features can support speech generation raises the possibility that they encode contrasts that are relevant to output-oriented phonological representation. In other words, the results suggest that FUL captures distinctions that are useful not only for analysing phonological systems, but also for generating spoken output. As a preliminary exploration of FUL-based TTS, the present findings are not intended to provide conclusive evidence for FUL theory, but rather to offer initial insights and directions for future research.

Our results also show that feature-based TTS can produce `accented' outputs that reflect the phonological characteristics of the training language when synthesising an unseen language. Rather than simply demonstrating multilingual transfer, this provides a computational analogue of the FUL hypothesis that cross-linguistic processing operates over abstract phonological representations rather than surface phonetic forms. In FUL accounts of loanword adaptation, borrowers infer the underlying phonological contrasts of the donor language and map them onto the contrastive feature system of the borrowing language, leading to language-specific realisations of the same input \citep{lahiriPertinacityLoanwordsSame2019}. Likewise, our model receives language-independent phonological feature representations as input but learns language-specific mappings from these features to acoustic output. Consequently, the same feature sequence can produce systematically different pronunciations depending on the language on which the model has been trained. Although this parallel should not be interpreted as a cognitive model of human speech processing, it suggests that feature-based representations may offer a useful computational framework for exploring how abstract phonological representations are transformed into language-specific speech output.

\subsection{Limitations and future directions}

As an initial exploration of FUL-based multilingual TTS, this study focused on assessing feasibility. Having established this, several directions for future research emerge, addressing model design and evaluation, as well as broader questions of controllability, data efficiency, and inclusivity in multilingual speech synthesis.

Some aspects of the current TTS pipeline can be improved to enhance output quality. For instance, the alignment tools used in this study were trained on ARPABET or Pinyin, whereas the input sequences were based on the more fine-grained SAMPA phone set. This mismatch may have introduced alignment errors, potentially leading to suboptimal synthesis quality.

Another direction is to integrate feature-based representations with more recent end-to-end TTS architectures, which may improve naturalness while maintaining interpretability and controllability, for example by enabling explicit control over phonological properties of the output. However, this would require developing major innovative methods to incorporate explicit phonological features into models that typically learn to generate speech directly from text and audio data, without relying on intermediate phoneme or feature representations.

Additionally, by examining typologically different languages, the present study revealed challenges associated with modelling lexical tone across languages. This suggests that prosodic features are not adequately captured in the current feature-based framework, which is expected given that the present study focused primarily on segmental FUL features and did not explicitly model prosodic structure using features. One possible future direction is to incorporate prosodic information, such as tone and intonation, into the input representation using more generalised encodings (e.g. High and Low), rather than relying on language-specific tonal or intonational markers, which may be less readily transferable across languages. \added{The results of the current study demonstrate that sparse segmental phonological features can be learned and transferred across languages to a considerable extent. This raises the possibility that prosodic features may also be represented within a similar feature-based framework, allowing models to learn prosodic contrasts in one language and transfer them to another. However, prosodic systems remain considerably less well understood than segmental systems, and it is still unclear which aspects are universal and which are language-specific. Developing transferable prosodic feature representations is therefore likely to be substantially more challenging. These challenges may become particularly evident in typologically diverse languages exhibiting phenomena such as complex or conditioned stress systems, language-specific syllabification preferences, or other prosodic structures that differ markedly from those represented in the training languages. Consequently, further descriptive, typological, and theoretical research is needed to understand the nature of prosodic features and how they can be incorporated into a feature-based framework for multilingual speech synthesis, ultimately contributing towards a more universal TTS input representation.}

More controlled experiments with balanced datasets and systematic variation in data size are needed to further evaluate the data efficiency of the models. For instance, future work could investigate how much/little data is required for a model to produce intelligible output in both seen and unseen languages, and how increases in training data for one language affect output quality in another. Such studies would provide clearer insight into the data efficiency of the approach and its potential applicability to low-resource languages. It would also be valuable to compare the proposed feature-based approach with baseline models using phone-based input in order to evaluate their relative advantages for multilingual TTS.

Finally, a key direction for future work is to evaluate the approach on genuinely low-resource languages in order to assess its practical utility and inclusivity. This would involve applying the method to languages with limited available data and resources, and evaluating the quality and intelligibility of the synthesised output in these languages. Such work would provide important insights into the potential of feature-based approaches for supporting multilingual TTS in under-represented languages and contributing to more equitable access to speech technology.

\subsection{Conclusion}

Taken together, these findings suggest that feature-based representations offer a promising, theory-informed direction for multilingual TTS. By providing a structured and potentially generalisable input format, feature-based representations may facilitate cross-lingual transfer while maintaining a degree of interpretability and controllability. At the same time, the present results highlight important limitations, particularly in modelling prosodic structure and language-specific contrasts. These findings point to the need for further work to refine feature-based approaches and to evaluate their applicability in more diverse and genuinely low-resource settings.

More broadly, these findings have implications for ongoing developments in speech and language technology. LLM-based AI systems can capture complex linguistic patterns from data, including patterns that may not be readily apparent to human analysis, but they typically require substantial amounts of training data. In contrast, humans can often infer linguistic rules from far smaller datasets. Today, most TTS development focuses on large-scale models, but we argue that there is still value in exploring more data-efficient, theory-informed approaches.

One lesson from the history of neural networks is inspiring here. Although neural networks were first proposed in the 1940s \citep{mcculloch1943,rosenblatt1958}, research continued only slowly in later decades. Their application to TTS, or AI-based research more broadly, was very limited until the 2010s when advances in deep learning, larger datasets, and more powerful computation empowered neural networks, sparking a new wave of AI development. Potentially and optimistically, combining theory-informed approaches with modern speech technology could lead to a similar breakthrough, potentially enabling more controllable, data-efficient “small language models” and making speech technology more inclusive for low-resource languages and varieties.

\section*{Acknowledgements}
The authors thank the participants for their time and effort in completing the listening tests. We are also truly grateful to the anonymous reviewers and the editor for their detailed and constructive feedback, which greatly improved the quality of the manuscript.

\section*{Author Contributions}

\textbf{Cong Zhang}: Conceptualisation, Methodology, Validation, Formal analysis, Investigation, Data curation, Writing - original draft, Writing - review \& editing, Visualisation, Project administration. \textbf{Huinan Zeng}: Methodology, Data curation, Writing - original draft, Writing - review \& editing, Visualisation. \textbf{Huang Liu}: Software, Writing - original draft. \textbf{Jiewen Zheng}: Conceptualisation, Methodology, Software, Writing - review \& editing.

\section*{Supplementary Materials}

\begin{itemize}
  \item Audio demos are publicly available: \url{https://congzhang365.github.io/feature_tts/}.
  \item Feature mapping, test utterance list, statistical analysis models and detailed results can be found in the OSF repository for this project: \url{https://osf.io/c78nz/}
\end{itemize}

\printbibliography[title={References}]

@inproceedings{maniati_cross-lingual_2021,
	title = {Cross-lingual low resource speaker adaptation using phonological features},
	volume = {5},
	isbn = {978-1-71383-690-2},
	doi = {10.21437/Interspeech.2021-327},
	pages = {3386--3390},
	booktitle = {Proceedings of the Annual Conference of the International Speech Communication Association, {INTERSPEECH}},
	author = {Maniati, Georgia and Ellinas, Nikolaos and Markopoulos, Konstantinos and Vamvoukakis, Georgios and Sung, June Sig and Park, Hyoungmin and Chalamandaris, Aimilios and Tsiakoulis, Pirros},
	year = {2021},
	eprinttype = {arxiv},
	eprint = {2111.09075},
	note = {{ISSN}: 19909772},
}

@inproceedings{wells_cross-lingual_2021,
	title = {Cross-lingual Transfer of Phonological Features for Low-resource Speech Synthesis},
	doi = {10.21437/ssw.2021-28},
	pages = {160--165},
	booktitle = {11th {ISCA} Speech Synthesis Workshop},
	author = {Wells, Dan and Richmond, Korin},
	year = {2021},
}

@article{zhang_hanyu_2009,
	title = {Hanyu Putonghua Jidu Yinbiao {SAMPA}-{SC} [{SAMPA}-{SC} for standard Chinese (Putonghua)]},
	journal = {Shengxue Xuebao},
	author = {Zhang, Jialu},
	year = {2009},
}

@article{gutkin_fonbund_2019,
	title = {Fonbund: A library for combining cross-lingual phonological segment data},
	pages = {2236--2240},
	journal = {{LREC} 2018 - 11th International Conference on Language Resources and Evaluation},
	author = {Gutkin, Alexander and Jansche, Martin and Merkulova, Tatiana},
	year = {2019},
	note = {{ISBN}: 9791095546009},
}

@article{staib_phonological_2020,
	title = {Phonological features for 0-shot multilingual speech synthesis},
	volume = {2020-Octob},
	issn = {19909772},
	doi = {10.21437/Interspeech.2020-1821},
	pages = {2942--2946},
	journal = {Proceedings of the Annual Conference of the International Speech Communication Association, {INTERSPEECH}},
	author = {Staib, Marlene and Teh, Tian Huey and Torresquintero, Alexandra and Ram Mohan, Devang S. and Foglianti, Lorenzo and Lenain, Raphael and Gao, Jiameng},
	year = {2020},
	eprinttype = {arxiv},
	eprint = {2008.04107},
}

@article{arora_phonological_2018,
	title = {Phonological feature-based speech recognition system for pronunciation training in non-native language learning},
	volume = {143},
	issn = {0001-4966},
	url = {http://dx.doi.org/10.1121/1.5017834},
	doi = {10.1121/1.5017834},
	pages = {98--108},
	number = {1},
	journal = {The Journal of the Acoustical Society of America},
	author = {Arora, Vipul and Lahiri, Aditi and Reetz, Henning},
	year = {2018},
	pmid = {29390749},
}

@article{ren_fastspeech_2019,
	title = {{FastSpeech}: Fast, robust and controllable text to speech},
	volume = {32},
	issn = {10495258},
	issue = {{NeurIPS}},
	journal = {Advances in Neural Information Processing Systems},
	author = {Ren, Yi and Ruan, Yangjun and Tan, Xu and Qin, Tao and Zhao, Sheng and Zhao, Zhou and Liu, Tie Yan},
	year = {2019},
	eprinttype = {arxiv},
	eprint = {1905.09263},
}

@article{chomsky_sound_1968,
	title = {The sound pattern of English.},
	author = {Chomsky, Noam and Halle, Morris},
	year = {1968},
	note = {Publisher: {ERIC}},
}

@incollection{lahiri_predicting_2018,
	location = {Berlin, Boston},
	title = {Predicting universal phonological contrasts},
	pages = {229--272},
	booktitle = {Phonological Typology},
	publisher = {De Gruyter Mouton},
	author = {Lahiri, Aditi},
	editor = {Hyman, Larry M and Plank, Frans},
	year = {2018},
	doi = {10.1515/9783110451931-007},

}

@incollection{LahiriReetz2002,
  author    = {Lahiri, Aditi and Reetz, Henning},
  title     = {Underspecified Recognition},
  booktitle = {Laboratory Phonology 7},
  editor    = {Gussenhoven, Carlos and Warner, Natasha},
  publisher = {Mouton de Gruyter},
  address   = {Berlin},
  year      = {2002},
  pages     = {637--676}
}

@incollection{ghiniPlaceArticulationFirst2012,
  title = {Place of Articulation First},
  booktitle = {Distinctive Feature Theory},
  author = {Ghini, Mirco},
  editor = {Hall, T Alan},
  year = {2012},
  pages = {147--176},
  publisher = {{De Gruyter Mouton}},
  doi = {10.1515/9783110886672.147}
}

@article{kennardNonesuchPhonemesLoanwords2020,
  title = {Nonesuch Phonemes in Loanwords},
  author = {Kennard, Holly J and Lahiri, Aditi},
  year = {2020},
  journal = {Linguistics},
  volume = {58},
  number = {1},
  pages = {83--108},
  publisher = {{De Gruyter}},
  issn = {0024-3949}
}

@incollection{kotzorSymmetryAsymmetryEvidence2017,
  title = {Symmetry or Asymmetry: {{Evidence}} for Underspecification in the Mental Lexicon},
  booktitle = {The {{Speech Processing Lexicon}}},
  author = {Kotzor, Sandra and Wetterlin, Allison and Lahiri, Aditi},
  editor = {Lahiri, Aditi and Kotzor, Sandra},
  year = {2017},
  pages = {85--106},
  publisher = {{De Gruyter Mouton}},
  doi = {10.1515/9783110422658-005}
}

@article{jakobsonPreliminariesSpeechAnalysis1952,
  title = {Preliminaries to Speech Analysis: {{The}} Distinctive Features and Their Correlates},
  author = {Jakobson, R. and Fant, G. and Halle, M.},
  year = {1952},
  journal = {Technical Report No. 13}
}

@article{wellsSAMPAComputerReadable1997,
  title = {{{SAMPA}} Computer Readable Phonetic Alphabet},
  author = {Wells, John C},
  year = {1997},
  journal = {Handbook of standards and resources for spoken language systems},
  volume = {4},
  pages = {684--732},
  publisher = {{Berlin and New York: Mouton de Gruyter. Part IV, section B}}
}

@article{Lee-Kim2014,
  title = {Revisiting {{Mandarin}} 'Apical Vowels': {{An}} Articulatory and Acoustic Study},
  author = {{Lee-Kim}, Sang Im},
  year = {2014},
  journal = {Journal of the International Phonetic Association},
  volume = {44},
  number = {3},
  pages = {261--282},
  issn = {14753502},
  doi = {10.1017/S0025100314000267},
  isbn = {0025100314000}
}

@book{Duanmu2000e,
  title = {The {{Phonology}} of {{Standard Chinese}}},
  author = {Duanmu, San},
  year = {2000},
  publisher = {{Oxford University Press}},
  isbn = {978-0-19-921578-2}
}

@inproceedings{poveyKaldiSpeechRecognition2011,
  title = {The {{Kaldi}} Speech Recognition Toolkit},
  booktitle = {{{IEEE}} 2011 Workshop on Automatic Speech Recognition and Understanding},
  author = {Povey, Daniel and Ghoshal, Arnab and Boulianne, Gilles and Burget, Lukas and Glembek, Ondrej and Goel, Nagendra and Hannemann, Mirko and Motlicek, Petr and Qian, Yanmin and Schwarz, Petr},
  year = {2011},
  number = {CONF},
  publisher = {{IEEE Signal Processing Society}}
}

@inproceedings{kumarMelGANGenerativeAdversarial2019,
  title = {{{MelGAN}}: {{Generative Adversarial Networks}} for {{Conditional Waveform Synthesis}}},
  shorttitle = {{{MelGAN}}},
  booktitle = {Advances in {{Neural Information Processing Systems}}},
  author = {Kumar, Kundan and Kumar, Rithesh and {de Boissiere}, Thibault and Gestin, Lucas and Teoh, Wei Zhen and Sotelo, Jose and {de Br{\'e}bisson}, Alexandre and Bengio, Yoshua and Courville, Aaron C},
  year = {2019},
  volume = {32},
  publisher = {{Curran Associates, Inc.}},
  urldate = {2023-08-03},
  langid = {english}
}

@article{stoetPsyToolkitNovelWebbased2017,
  
  title = {{PsyToolkit}: {A} Novel Web-Based Method for Running Online Questionnaires and Reaction-Time Experiments},
  author = {Stoet, Gijsbert},
  year = {2017},
  journal = {Teaching of Psychology},
  volume = {44},
  number = {1},
  pages = {24--31},
  publisher = {{Sage Publications Sage CA: Los Angeles, CA}},
  issn = {0098-6283}
}

@article{stoetPsyToolkitSoftwarePackage2010,
  title = {{{PsyToolkit}}: {{A}} Software Package for Programming Psychological Experiments Using {{Linux}}},
  author = {Stoet, Gijsbert},
  year = {2010},
  journal = {Behavior Research Methods},
  volume = {42},
  number = {4},
  pages = {1096--1104},
  issn = {1554-3528},
  doi = {10.3758/BRM.42.4.1096}
}

@misc{zhangPhonologicalFeatureMapping2021,
  title = {Phonological Feature Mapping for {{FeatureTTS}}},
  author = {Zhang, Cong and Zeng, Huinan},
  year = {2021},
  month = oct,
  doi = {10.5281/ZENODO.5553685},
  urldate = {2021-10-07},
  copyright = {All rights reserved}
}

@inproceedings{zhuPhonetoAudioAlignmentText2022,
  title = {Phone-to-{{Audio Alignment}} without {{Text}}: {{A Semi-Supervised Approach}}},
  shorttitle = {Phone-to-{{Audio Alignment}} without {{Text}}},
  booktitle = {{{ICASSP}} 2022 - 2022 {{IEEE International Conference}} on {{Acoustics}}, {{Speech}} and {{Signal Processing}} ({{ICASSP}})},
  author = {Zhu, Jian and Zhang, Cong and Jurgens, David},
  year = {2022},
  month = may,
  pages = {8167--8171},
  issn = {2379-190X},
  doi = {10.1109/ICASSP43922.2022.9746112},
  urldate = {2024-08-13}
}

@article{Lahiri2010,
  title = {Distinctive Features: {{Phonological}} Underspecification in Representation and Processing},
  author = {Lahiri, Aditi and Reetz, Henning},
  year = {2010},
  month = jan,
  journal = {Journal of Phonetics},
  volume = {38},
  number = {1},
  pages = {44--59},
  issn = {00954470},
  doi = {10.1016/j.wocn.2010.01.002},
  urldate = {2014-04-30}
}

@article{mcculloch1943,
  author    = {McCulloch, W. S. and Pitts, W.},
  title     = {A logical calculus of the ideas immanent in nervous activity},
  journal   = {{The Bulletin of Mathematical Biophysics}},
  year      = {1943},
  volume    = {5},
  number    = {4},
  pages     = {115--133}
}

@article{rosenblatt1958,
  author    = {Rosenblatt, Frank},
  title     = {The perceptron: A probabilistic model for information storage and organization in the brain},
  journal   = {Psychological Review},
  year      = {1958},
  volume    = {65},
  number    = {6},
  pages     = {386--408}
}

@article{wang2023neural,
  title={Neural codec language models are zero-shot text to speech synthesizers},
  author={Wang, Chengyi and Chen, Sanyuan and Wu, Yu and Zhang, Ziqiang and Zhou, Long and Liu, Shujie and Chen, Zhuo and Liu, Yanqing and Wang, Huaming and Li, Jinyu and others},
  journal={arXiv preprint arXiv:2301.02111},
  year={2023}
}

@inproceedings{taannander2024beyond,
  title={Beyond graphemes and phonemes: continuous phonological features in neural text-to-speech synthesis},
  author={T{\aa}nnander, Christina and Mehta, Shivam and Beskow, Jonas and Edlund, Jens},
  journal={Proc. Interspeech 2024},
  pages={2815--2819},
  year={2024},
  doi={doi: 10.21437/Interspeech.2024-1565}
}

@inproceedings{louw2023cross,
  title={Cross-lingual transfer using phonological features for resource-scarce text-to-speech},
  author={Louw, Johannes Abraham},
  booktitle={12th Speech Synthesis Workshop (SSW) 2023},
  year={2023}
}

@article{huang2025step,
  title={Step-audio: Unified understanding and generation in intelligent speech interaction},
  author={Huang, Ailin and Wu, Boyong and Wang, Bruce and Yan, Chao and Hu, Chen and Feng, Chengli and Tian, Fei and Shen, Feiyu and Li, Jingbei and Chen, Mingrui and others},
  journal={arXiv preprint arXiv:2502.11946},
  year={2025}
}

@bachelorsthesis{naslund2024hypernasality,
  author       = {Alexander Näslund},
  title        = {Simulating hypernasality with phonological features in Swedish TTS},
  school       = {Department of Linguistics, Bachelor’s Programme in Linguistics, LIN622},
  type         = {Bachelor's Thesis},
  year         = {2024},
  semester     = {Spring},
  credits      = {15 ECTS},
  supervisors  = {Marcin Włodarczak and Christina Tånnander},
  note         = {Swedish title: Simulerad hypernasalitet med fonologiska särdrag i svensk TTS}
}

@article{klatt1987review,
  title={Review of text-to-speech conversion for English},
  author={Klatt, Dennis H.},
  journal={The Journal of the Acoustical Society of America},
  volume={82},
  number={3},
  pages={737--793},
  year={1987},
  publisher={Acoustical Society of America}
}

@inproceedings{tokuda2000speech,
  title={Speech parameter generation algorithms for HMM-based speech synthesis},
  author={Tokuda, Keiichi and Yoshimura, Takayoshi and Masuko, Takashi and Kobayashi, Takao and Kitamura, Tadashi},
  booktitle={ICASSP},
  volume={3},
  pages={1315--1318},
  year={2000},
  organization={IEEE}
}

@inproceedings{zen2013statistical,
  title={Statistical parametric speech synthesis using deep neural networks},
  author={Zen, Heiga and Senior, Andrew and Schuster, Mike},
  booktitle={ICASSP},
  pages={7962--7966},
  year={2013},
  organization={IEEE}
}

@inproceedings{wang2017tacotron,
  title={Tacotron: Towards end-to-end speech synthesis},
  author={Wang, Yuxuan and Skerry-Ryan, RJ and Stanton, Daisy and Wu, Yonghui and Weiss, Ron J. and Jaitly, Navdeep and Yang, Zongheng and Xiao, Ying and Chen, Zhifeng and Bengio, Samy and Le, Quoc V. and Agiomyrgiannakis, Yannis and Clark, Rob and Saurous, Rif A.},
  booktitle={Interspeech},
  pages={4006--4010},
  year={2017}
}

@inproceedings{shen2018natural,
  title={Natural TTS synthesis by conditioning WaveNet on mel spectrogram predictions},
  author={Shen, Jonathan and Pang, Ruoming and Weiss, Ron J. and Schuster, Mike and Jaitly, Navdeep and Yang, Zongheng and Chen, Zhifeng and Zhang, Yu and Wang, Yuxuan and Skerry-Ryan, RJ and Saurous, Rif A. and Agiomyrgiannakis, Yannis and Wu, Yonghui},
  booktitle={ICASSP},
  pages={4779--4783},
  year={2018},
  organization={IEEE}
}

@inproceedings{ren2021fastspeech2,
  title={FastSpeech 2: Fast and high-quality end-to-end text to speech},
  author={Ren, Yi and Hu, Chenxu and Tan, Xu and Qin, Tao and Zhao, Sheng and Zhao, Zhou and Liu, Tie-Yan},
  booktitle={ICLR},
  year={2021}
}

@article{Hauser2023,
author = {Ivy Hauser},
title ={Differential Cue Weighting in Mandarin Sibilant Production},

journal = {Language and Speech},
volume = {66},
number = {4},
pages = {1056-1090},
year = {2023},
doi = {10.1177/00238309231152495},
URL = {https://doi.org/10.1177/00238309231152495}
}

@Manual{ordinal,
    title = {ordinal---Regression Models for Ordinal Data},
    author = {Rune H. B. Christensen},
    year = {2023},
    note = {R package version 2023.12-4.1},
    url = {https://CRAN.R-project.org/package=ordinal},
  }

@Manual{R-base,
  title        = {R: A Language and Environment for Statistical Computing},
  author       = {{R Core Team}},
  organization = {R Foundation for Statistical Computing},
  address      = {Vienna, Austria},
  year         = {2024},
  note         = {Version 4.4.2},
  url          = {https://www.R-project.org/}
}

@Manual{emmeans,
    title = {emmeans: Estimated Marginal Means, aka Least-Squares Means},
    author = {Russell V. Lenth},
    year = {2025},
    note = {R package version 1.10.7},
    url = {https://CRAN.R-project.org/package=emmeans},
  }

@Article{lme4,
    title = {Fitting Linear Mixed-Effects Models Using {lme4}},
    author = {Douglas Bates and Martin M{\"a}chler and Ben Bolker and Steve Walker},
    journal = {Journal of Statistical Software},
    year = {2015},
    volume = {67},
    number = {1},
    pages = {1--48},
    doi = {10.18637/jss.v067.i01},
  }

@inproceedings{liAcousticAnalysisMandarin2015,
  title = {Acoustic Analysis of {{Mandarin}} Affricates},
  booktitle = {Interspeech 2015},
  author = {Li, Shanpeng and Gu, Wentao},
  year = 2015,
  pages = {1680--1684},
  doi = {10.21437/Interspeech.2015-387},


}

@article{maniwa2009,
  title = {Acoustic Characteristics of Clearly Spoken {{English}} Fricatives},
  author = {Maniwa, Kazumi and Jongman, Allard and Wade, Travis},
  year = 2009,
  month = jun,
  journal = {The Journal of the Acoustical Society of America},
  volume = {125},
  number = {6},
  eprint = {https://pubs.aip.org/asa/jasa/article-pdf/125/6/3962/13363044/3962\_1\_online.pdf},
  pages = {3962--3973},
  issn = {0001-4966},
  doi = {10.1121/1.2990715}
}

@incollection{lahiriPertinacityLoanwordsSame2019,
  author    = {Lahiri, Aditi and Kennard, Holly},
  title     = {Pertinacity in loanwords: Same underlying systems, different outputs},
  booktitle = {Historical Linguistics 2015: Selected Papers from the 22nd International Conference on Historical Linguistics, Naples, 27--31 July 2015},
  editor    = {Cennamo, Michela and Fabrizio, Claudia},
  series    = {Current Issues in Linguistic Theory},
  volume    = {348},
  pages     = {57--74},
  publisher = {John Benjamins},
  address   = {Amsterdam},
  year      = {2019},
  doi       = {10.1075/cilt.348.03lah},
  url       = {https://doi.org/10.1075/cilt.348.03lah}
}

@article{althausCoronalUnderspecificationEmerging2024,
  title = {Coronal Underspecification as an Emerging Property in the Development of Speech Processing},
  author = {Althaus, Nadja and Lahiri, Aditi and Plunkett, Kim},
  year = 2024,
  journal = {Journal of Experimental Psychology: Learning, Memory, and Cognition},
  volume = {50},
  number = {12},
  pages = {1932--1953},
  publisher = {American Psychological Association},
  address = {US},
  issn = {1939-1285},
  doi = {10.1037/xlm0001367},

}

@inproceedings{sanchez2025,
  author    = {Ariadna Sanchez and Simon King},
  title     = {Can We Reconstruct a Dysarthric Voice with the Large Speech Model Parler TTS?},
  booktitle = {Proceedings of Interspeech 2025},
  pages     = {4138--4142},
  year      = {2025},
  publisher = {ISCA},
  doi       = {10.21437/Interspeech.2025-2679}
}

\end{document}